\documentclass[11pt]{article}
\ifdefined\pdfinfoomitdate\pdfinfoomitdate=1\fi
\ifdefined\pdftrailerid\pdftrailerid{}\fi

\usepackage[preprint]{acl}

\usepackage{times}
\usepackage{latexsym}
\usepackage[T1]{fontenc}
\usepackage[utf8]{inputenc}
\usepackage{microtype}
\usepackage{inconsolata}
\usepackage{graphicx}
\usepackage{booktabs}
\usepackage{adjustbox}  
\usepackage{xcolor}
\usepackage{pgfplots}
\pgfplotsset{compat=1.16}

\newcommand{\cndcode}[1]{\texttt{#1}}
\makeatletter
\edef\@cndus{\string_}
\DeclareRobustCommand{\cnd}[1]{\begingroup\let\_\@cndus\edef\@cndtmp{#1}\expandafter\@cndout\expandafter{\@cndtmp}\endgroup}
\newcommand{\@cndout}[1]{\ifcsname cnd@#1\endcsname\csname cnd@#1\endcsname\else\texttt{#1}\fi}
\@namedef{cnd@none}{no count}
\@namedef{cnd@prose_up_to_k}{prose ``up to $k$''}
\@namedef{cnd@prose_exactly_k}{prose ``exactly $k$''}
\@namedef{cnd@max_k_shown}{shown maximum}
\@namedef{cnd@min_k_shown}{shown minimum}
\@namedef{cnd@min_k_shown_fewer_ok}{``fewer is fine''}
\@namedef{cnd@fewer_ok}{``fewer is fine''}
\@namedef{cnd@min_k_enforced}{enforced minimum}
\@namedef{cnd@min_k_escape}{N/A escape}
\makeatother
\newcommand{\pre}{$^{\dagger}$}


\title{Padding to $k$: A Minimum Count in the Schema Makes\\Open-Weight Models Pad Their Extraction Lists}

\author{Abhishek Barali \\ St.\ Xavier's College \\ Maitighar, Kathmandu \\ Tribhuvan University, Nepal \\ \texttt{abhibidut@gmail.com} \\ ORCID \href{https://orcid.org/0009-0007-3317-8437}{0009-0007-3317-8437} \And
  Aayush Paudel \\ St.\ Xavier's College \\ Maitighar, Kathmandu \\ Tribhuvan University, Nepal \\ \texttt{aayushpaudel09@gmail.com} \\ ORCID \href{https://orcid.org/0009-0001-0962-2821}{0009-0001-0962-2821}}

\begin{document}
\makeatletter
\hypersetup{pdftitle={Padding to k: A Minimum Count in the Schema Makes Open-Weight Models Pad Their Extraction Lists},
  pdfauthor={\ifacl@anonymize Anonymous\else Abhishek Barali, Aayush Paudel\fi}}
\makeatother
\maketitle

\begin{abstract}
Extraction pipelines ask language models for JSON that follows a schema, and the schema can set a minimum number of list entries.
We ask what models do when the document holds fewer items than that minimum.
Under a preregistered plan, we test eight open-weight models on 558 real documents from nine annotated datasets, with six more models on hosted services as a reference.
Three findings stand out.
First, models treat the count as a target: a minimum shown in the schema, but not enforced, raises the share of lists longer than the true count from 34.5\% to 55.4\%.
Second, the padding is hard to catch, because most added entries are real text taken from the document; about half of them pass a check that each entry occurs in the source.
Third, common fixes help only partly: a line saying fewer items are fine removes the effect on average but misses more true items, a count in prose (``up to $k$'') also raises over-counting, an ``N/A'' option under an enforced minimum still leaves half the lists padded, and re-asking with the validation error makes padding far more common.
We release the code, answer lists and run records at \url{https://github.com/AbhishekBarali/padding-to-k}.
\end{abstract}

\section{Introduction}
\label{sec:intro}

\begin{figure}[t]
  \centering
  \fbox{\begin{minipage}{0.92\columnwidth}
    \small
    \textbf{Document} (one email; its answer list has $n = 1$ request):\\
    \textit{Subject: Cost Allocations}\\ \textit{In connection with the Legal costs allocations for 2002, what percentage of your time do you think you would spend on EGM matters? The accounting group who is asking for this information realizes that this is a guess on your part. Thanks, Alan}\\[4pt]
    \textbf{Prompt}: ``List all the requests.'', with a JSON Schema whose \texttt{requests} array has \texttt{"minItems": 3}, shown in the prompt and not enforced.\\[4pt]
    \textbf{Output} (Qwen3.5-9B, full run):\\
    1. ``In connection with the Legal costs allocations for 2002, what percentage of your time do you think you would spend on EGM matters?''\\
    2. ``The accounting group who is asking for this information realizes that this is a guess on your part.''\\
    3. ``Thanks, Alan''
  \end{minipage}}
  \caption{An email with one request, a prompt that asks for at least three, and the list Qwen3.5-9B returned. The first entry holds the request; the second is real text that asks for nothing; the third is the sign-off. Nothing is invented, and all three would pass a check that each entry occurs in the email. The example was picked to illustrate the pattern; it is not a random draw.}
  \label{fig:example}
\end{figure}

Extraction systems ask language models for structured output: a JSON object that follows a JSON Schema.
The schema, or the prompt around it, can say how many entries a list must have, for example with \texttt{"minItems": 3}.
That is harmless when the document holds at least that many items.
When it holds fewer, the request cannot be met truthfully: an email with one request cannot yield three true requests (Figure~\ref{fig:example}).
Real schemas set list minimums: of the 9,542 real-world schemas in JSONSchemaBench \citep{geng2025jsonschemabench}, 7.7\% put a \texttt{minItems} of at least 1 on a list of strings or objects, which an empty document cannot meet; minimums of 2 or more are rarer (0.3\%; post hoc, Appendix~\ref{app:posthoc}).

We fix what a correct output is in advance (Sections~\ref{sec:scoring} and~\ref{sec:prereg}).
When the document holds $n < k$ true items, a correct output lists the $n$ items and stops, or marks the list as short with a placeholder, an ``N/A'' where one is offered, or a note.
Padding the list with entries that are not true items is the failure, whether the padding is invented or taken from the document.
Section~\ref{sec:scoring} gives the reasons for this reading, and Section~\ref{sec:discussion} the evidence that the minimum, not the schema, causes the padding.

\begin{figure}[t]
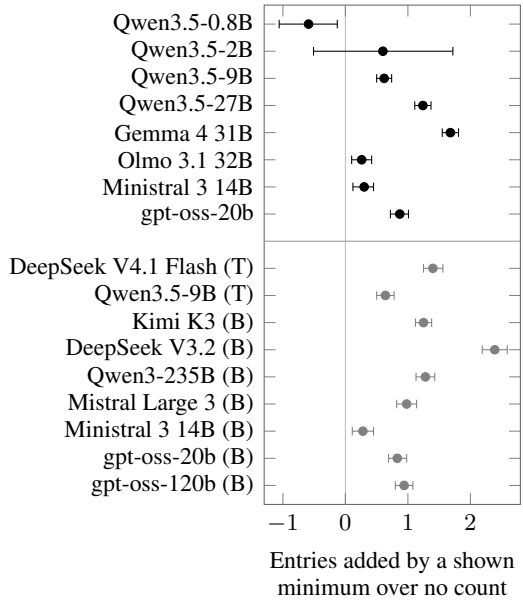

  \centering
  \input{generated/fig_c2_ticks.tex}
  \begin{tikzpicture}
    \begin{axis}[scale only axis, width=0.44\columnwidth, height=6.6cm, xmin=-1.3, xmax=2.8,
                 xlabel={Entries added by a shown\\minimum over no count}, xlabel style={align=center},
                 padkCtwoTicks,
                 y tick label style={font=\footnotesize}, x tick label style={font=\footnotesize},
                 label style={font=\small}, ymin=0.3, ymax=18.7,
                 extra x ticks={0}, extra x tick style={grid=major, tick label style={opacity=0}},
                 axis line style={black!60}]
      \input{generated/fig_c2_points.tex}
      \draw[black!40] (axis cs:-1.3,10) -- (axis cs:2.8,10);
    \end{axis}
  \end{tikzpicture}
  \caption{The main contrast per model: substantive entries under a \texttt{minItems} shown in the schema minus entries with no count, same document, documents with $n < k$, $k = 3$ and 5, with 95\% group-bootstrap intervals. Black: the eight GPU models. Grey, below the line: hosted reference models on Together (T) and Bedrock (B), never pooled with the GPU models.}
  \label{fig:headline}
\end{figure}

Stated counts and required fields are known to push models toward content that is not there \citep{yang2026prompt,ivgi2025loops,usman2026phantomfill}.
The closest study, PhantomFill \citep{usman2026phantomfill}, puts minimum-count arrays in JSON schemas, but its inputs hold no true items by construction and it does not vary the count.
No study we know of varies a schema count against the true count on real documents, compares a minimum with a maximum, or measures how much of the added content comes from the document itself.
We test eight open-weight models on one GPU, and six more on two hosted services, on 558 documents that hold 0 to 6 checked items.
Each model is asked for 1, 3 or 5 items, with the count in prose, in a schema shown in the prompt, or in a schema the decoder enforces.
The main measure is the length change: for the same document and model, the number of entries under a shown minimum minus the number with no count (Figure~\ref{fig:headline}).

We find three things (Section~\ref{sec:results} marks which results are registered tests and which are exploratory or post hoc).

\textbf{Models treat the count as a target.}
A minimum shown in the schema, but not enforced, raises the share of lists longer than the true count from 34.5\% to 55.4\%: on the same documents, about one more list in five runs past the true count.
On the documents a model gets exactly right with no count, a shown minimum still makes about a third of its lists pad (35.2\%; post hoc).
Six of the eight GPU models and all nine hosted runs lengthen their lists; the exceptions are the two smallest models, one of which often writes no readable list (Section~\ref{sec:main}).
Registered controls tie the effect to the slots the document cannot fill: a minimum adds nothing on documents that already hold enough items, and a shown maximum barely changes over-count (Section~\ref{sec:named}).

\textbf{The padding is hard to catch.}
Most added entries are real text taken from the document, like the sign-off in Figure~\ref{fig:example}.
They parse, meet the schema, and about half pass a check that each entry occurs in the source (Section~\ref{sec:what}).

\textbf{Common fixes help only partly.}
A line saying fewer items are fine removes the effect on average but misses more true items.
An ``N/A'' option under an enforced minimum lowers padding only from 90.1\% to 51.0\%.
``Up to $k$'' in prose also raises over-counting, from 34.5\% to 42.7\%.
Re-asking with the bare validation error, as schema-validating libraries do, made padding far more common in every run tested, from 10.0\% to 69.0\% for Qwen3.5-9B (Section~\ref{sec:exploratory}).

Our contributions:
\begin{itemize}
  \setlength\itemsep{0pt}
  \item 558 real documents from nine annotated sets with answer lists checked by both authors, released with the code and run records (Section~\ref{sec:data}, Appendix~\ref{app:data}).
  \item A preregistered, per-model measurement of count pressure on eight open-weight models, with controls, and six hosted models as a reference (Section~\ref{sec:results}).
  \item Evidence that the padding is mostly real document text, with practical guidance for schema authors (Table~\ref{tab:guidance}).
\end{itemize}

\section{Related work}
\label{sec:related}

\paragraph{Forms and counts as pressure.}
The closest study to ours, PhantomFill \citep{usman2026phantomfill}, builds inputs that hold no answer by construction, such as a post with engagement counts but no reply text, and finds that required JSON fields drive fabrication to 100\% in ten of thirteen models and that open-weight models ignore an offered ``insufficient evidence'' value, even under enforced decoding.
Its schemas include minimum-count arrays, which GPT-5.5 filled with invented content in 3 of 20 trials.
Because its inputs hold nothing true, every filled slot is invented there.
Our documents hold some true items, so a padded list can also be filled with real text, and we vary $k$ against $n$.
\citet{yang2026prompt} states an error count in prose for CoNLL-2014 passages of learner English: moving it from two below the true count to two above moves the reported count of four GPT and Claude models by 82 to 94\% of that gap, with one case of GPT-4o inventing errors to reach the stated number, while Gemini models barely move.
\citet{ivgi2025loops} ask, from memory, for 25 facts where at most 10 exist, and lists move from correct facts to invented ones and then to repetitions.
Neither uses a schema or compares a minimum with a maximum.
Instruction-following benchmarks score such counts as goals to meet, for example ``mention the keyword of AI at least 3 times'' in IFEval \citep{zhou2023instruction}; on a document with fewer items than asked, meeting the count and staying true to the document conflict.

\paragraph{What extraction errors are made of.}
Two audits find that most false positives of model-based extraction are real text, not invention.
In \citet{han2023empirical}, spans the test set does not annotate are 11.7 to 48.5\% of GPT-4's errors on four information-extraction tasks, while spans absent from the input are 0 to 4.3\%.
\citet{qian2026unreliable} audits entity extraction from threat-intelligence reports under a preregistered protocol and estimates that 65.6\% of false positives are background mentions, correct strings from the wrong context, against 1.2\% with no support in the report; a hard cap of three items raised the precision of one entity type from 0.060 to 0.381.
Neither sets a minimum count.
We find the same kind of error under count pressure, where the minimum itself asks for it.

\paragraph{Prompt against decoder.}
Output formats change what models produce: JSON and similar formats lower reasoning accuracy, more so the stricter the format \citep{tam2024speak}.
\citet{lee2026format} separate a format requested in the prompt from a format enforced by the decoder and find that the prompt causes most of the accuracy loss on reasoning tasks.
Our shown and enforced conditions make the same split for a count: the prompt and schema are identical, and only the decoder differs.
JSONSchemaBench tests six constrained-decoding frameworks, XGrammar among them, on 10K real-world schemas for efficiency, coverage of constraint types and output quality \citep{geng2025jsonschemabench}; it asks whether the decoder meets the schema, and we ask what the model puts in the slots the schema forces.
Grammar-constrained decoding can also distort the model's distribution, giving grammatical outputs at likelihoods out of proportion to the model's own \citep{park2026grammar}, one reason an enforced minimum need not behave like a shown one.

\paragraph{Abstention.}
Saying that a list is short is a form of abstention, which is studied mostly as declining to answer questions that are underspecified or unanswerable \citep{wen2025know}.
AbstentionBench finds that reasoning fine-tuning lowers abstention by 24\% on average \citep{kirichenko2025abstentionbench}, in line with our thinking side test, where thinking raised padding for two of three models.
Our ``fewer is fine'' line and N/A escape are abstention options inside a schema.

\section{Data and design}
\label{sec:setup}

\paragraph{Terms.}
A document has $n$ true items, the entries of its answer list.
A request asks for $k$ items; when $k > n$ it leaves $k - n$ \emph{forced slots} that no true item can fill.
An output's \emph{substantive entries} are its entries that are neither placeholders (``N/A'', ``none found'') nor string-level repeats of an earlier entry.
The output \emph{over-counts} when it has more substantive entries than $n$; its \emph{excess} is how many more.
An \emph{extra entry} is a substantive entry that the item matcher (Section~\ref{sec:scoring}) does not match to a distinct true item.
Extra entries that the matcher finds in the document, verbatim or closely paraphrased, but that are not items of the kind asked for are \emph{grounded padding} (the matcher's ``wrong kind'' label).
A \emph{source check} keeps only the entries that occur in the source document.

\subsection{Documents and answer lists}
\label{sec:data}

The 558 documents come from nine sets drawn from eight published datasets with human annotations of the items we ask for: action items in AMI meetings \citep{carletta2005ami}, requests in MailEx emails \citep{srivastava2023mailex}, items on CORD receipts \citep{park2019cord}, skills and, separately, knowledge (tools) in SkillSpan job ads \citep{zhang2022skillspan}, medications in MACCROBAT case reports \citep{caufield2019maccrobat}, features in ROAST product reviews \citep{roast}, people in Few-NERD Wikipedia sentences \citep{ding2021fewnerd}, and direct speech in PDNC novels \citep{vishnubhotla2022pdnc}.
Each set has 15 documents for each $n$ from 0 to 3 (two cells hold 11 and 12; receipts and reviews have no $n = 0$ documents), and six sets add documents with $n = 4$ to 6 for asking fewer than exist (per-set totals in Table~\ref{tab:keysource}).
Long originals are cut into windows on sentence or speaker-turn boundaries, so each window holds a known number of whole items.

The answer lists start from the datasets' own annotations, which we cleaned under written counting rules that fix one reading for ambiguous cases, such as a request phrased twice.
We set the rules, including the five calls that move the most documents; an AI assistant applied them to all 1,446 candidate lists under our direction, keeping the annotators' original beside every change; author B checked every resulting decision by hand; and we settled every case the rules left open.
Documents were dropped for recorded reasons, mostly that they had no single right count; the 558 study documents fill the design's cells, and the reserve documents served only to develop the item matcher and the placeholder rules.

The lists pass the registered blind check: each author counted 40 documents from eight sets alone, without seeing the lists, and matched the list's count on 35 (87.5\%) and 38 (95.0\%), above the threshold of 85\% (Gwet's AC1 between the authors 0.851).
The reviews set, whose text may not be shared, was counted blind in a separate check of five documents: author A matched the list on all five and author B on four (Appendix~\ref{app:key}).
Job-ad skills is the least certain list (2 and 3 of 5 matches).
The cleaned count equals the annotators' count on 423 of the 558 documents (76\%), and the confirmatory contrasts are also given with the annotators' counts and per set (Appendices~\ref{app:key} and~\ref{app:perset}).

\subsection{Conditions}
\label{sec:conditions}

Every prompt is one user message: an instruction, the JSON schema pasted the way LangChain's \texttt{JsonOutputParser} pastes it, then the document.
The schema describes the item in one line (each set's line is quoted with its example in Appendix~\ref{app:posthoc}).
Table~\ref{tab:conditions} lists the eight conditions; each runs at $k = 1$, 3 and 5 (the no-count prompt does not depend on $k$ and is sent once), with all five wordings of each prose condition.
The schema conditions keep one instruction (``List all the requests.'' for emails) and differ only in a \texttt{minItems} or \texttt{maxItems} keyword, an added line or the decoder; the shown and the enforced minimum send the same prompt.
The ``fewer is fine'' line is ``If the document contains fewer than $k$, list all available ones.'', adapted from a real extraction prompt in ScrapeGraphAI-100k \citep{brach2026scrapegraphai}.
The prose conditions replace the instruction with a count wording modelled on prompts from the same dataset, such as ``Extract up to $k$ requests from this document.'' or ``Return the top $k$ requests.'', and keep the schema without a count (Table~\ref{tab:wordings}), so they differ from no count in the instruction as well as the count.

\begin{table}[t]
  \centering
  \small
  \setlength{\tabcolsep}{3pt}
  \begin{tabular}{@{}lll@{}}
    \toprule
    Condition & Count stated & Decoder \\
    \midrule
    \cnd{none} & no count & free \\
    \cnd{prose\_up\_to\_k} & ``up to $k$'', 5 wordings & free \\
    \cnd{prose\_exactly\_k} & ``exactly $k$'', 5 wordings & free \\
    \cnd{max\_k\_shown} & \texttt{maxItems} in the schema & free \\
    \cnd{min\_k\_shown} & \texttt{minItems} in the schema & free \\
    \cnd{min\_k\_shown\_fewer\_ok} & as above, plus ``fewer is fine'' & free \\
    \cnd{min\_k\_enforced} & \texttt{minItems} in the schema & enforced \\
    \cnd{min\_k\_escape} & as above, plus ``N/A'' allowed & enforced \\
    \bottomrule
  \end{tabular}
  \caption{Conditions. The first six leave the decoder free; the Bedrock models run only these. In the N/A escape the instruction adds ``If the information is not available, use N/A.'' and the schema allows it. In the code, in this order: \cndcode{none}, \cndcode{prose\_up\_to\_k}, \cndcode{prose\_exactly\_k}, \cndcode{max\_k\_shown}, \cndcode{min\_k\_shown}, \cndcode{min\_k\_shown\_fewer\_ok}, \cndcode{min\_k\_enforced}, \cndcode{min\_k\_escape}.}
  \label{tab:conditions}
\end{table}

\subsection{Models}
\label{sec:models}

Eight open-weight models from five families run on one GPU: Qwen3.5-0.8B, -2B, -9B and -27B, Gemma 4 31B, Olmo 3.1 32B, Ministral 3 14B and gpt-oss-20b.
Each is pinned to a Hugging Face revision and served by vLLM v0.31.0 \citep{kwon2023vllm} with XGrammar 0.2.7 \citep{dong2025xgrammar}, in BF16 (gpt-oss-20b in its native MXFP4), with greedy decoding (temperature 0, seed 0) and thinking off; gpt-oss cannot turn reasoning off and runs at low effort.
We test open-weight models only.
Closed APIs, which carry much production extraction, are the main gap, and they may behave differently: in PhantomFill, GPT-5.5 filled a minimum-count array with invented content in only 3 of 20 trials \citep{usman2026phantomfill}.
As a reference, never pooled with the GPU models, six more open-weight models run on hosted services: DeepSeek V4.1 Flash (thinking off) on Together AI in every condition, and Kimi K3 (thinking off), DeepSeek V3.2, Qwen3-235B, Mistral Large 3 and gpt-oss-120b on Amazon Bedrock in the unenforced conditions only, because Bedrock cannot enforce a minimum above 1.
Three GPU models also run hosted (Qwen3.5-9B on Together, Ministral 3 14B and gpt-oss-20b on Bedrock) to test whether the serving stack matters.
No Claude model is run, because a Claude model helped prepare the answer lists.

\subsection{Scoring}
\label{sec:scoring}

\paragraph{Which instruction a correct output follows.}
When $n < k$, ``List all the requests.'' and the schema's \texttt{minItems} cannot both be met with true entries, so the study has to say which one a correct output follows.
The registered plan chose the instruction, for three reasons.
The instruction is about the document in hand, while a schema is written once for a whole pipeline, before any document is seen; an instruction followed by the schema is the layout LangChain's \texttt{JsonOutputParser} is built for.
A short list is correct data, and a placeholder or a note says that items are missing, while a padded list passes every format check and is wrong.
And a reading under which the minimum wins would accept any filler, invented entries included.
The main measure does not depend on this choice: the length change counts how far the stated number moves the list under either reading, and only calling the padding a failure rests on it.

Code reads every output through one parse ladder (strict JSON, a fenced block, the first object holding the list, a bare list, a cut-off list); a copy of the schema is never taken as the answer, so an output that only fills in a schema copy is unreadable.
The main measure is \emph{length change}: the number of substantive entries under a condition minus the number under its comparison condition, for the same document and model (for contrast 2, a shown minimum against no count).
It needs the true count $n$ only to choose the documents with $n < k$, not item-level matching, and because the comparison is paired, the extraction errors a model already makes with no count drop out of it.
\emph{Over-count} is the secondary measure.
Every output on a document with $n < k$ is also sorted as padded (it over-counts, even if it also flags the shortfall), says it is short (a placeholder, an ``N/A'' or a note), a short list with no note (at most $n$ substantive entries), or unreadable.
An item matcher labels every substantive entry as a true item, a split of one, grounded padding or invented, and gives the share of true items that no entry matches (\emph{left out}, on documents with $n \geq 1$).
The matcher was developed on reserve documents and frozen with the registered plan.
Against 400 entries that both authors labelled independently, it is right on 98\% of its true-item (or split) calls, 85\% of its wrong-kind calls and 69\% of its invented calls, and only 5 of the 400 entries were true items missing from the answer lists (Appendix~\ref{app:checks}).
Its invented shares are therefore its least firm numbers; every number that rests on it is marked \pre{}.
No language model judges any headline number.

\subsection{Preregistration and statistics}
\label{sec:prereg}

We froze a first plan, ran a pilot on 60 documents with three GPU and four hosted models, and registered the final plan on OSF (\url{https://osf.io/e4r9f}) before the full run.
Five contrasts are confirmatory and corrected together with Holm's method: (1) a larger minimum, a shown minimum at $k = 5$ against $k = 3$ on the same document with $n \leq 2$; (2) a shown minimum against no count, the main measure; (3) a shown minimum against a shown maximum; (4) ``exactly $k$'' against ``up to $k$''; and (5) the N/A escape against the plain enforced minimum, in excess entries per forced slot.
All are on documents with $n < k$ at $k = 3$ and 5.
For each, the one primary test is the 95\% interval of the mean paired difference over documents and the eight GPU models, from 10,000 bootstrap draws over document groups (an original meeting, email folder, job ad, case report or novel chapter) within set; a mixed model (linear for contrasts 1 to 4, Poisson for contrast 5) and each model's own estimate are reported next to it, per model first as the protocol asks.
Four named tests, three controls and an empty-document test, are reported whatever their result, outside the Holm family (Section~\ref{sec:named}).
The hosted runs, the per-set and per-wording results and the re-ask and thinking side tests are exploratory, and analyses added after the run are marked post hoc.
The interval describes these eight models on these documents; the per-model, family-level (Appendix~\ref{app:generality}) and hosted results speak to other models.

\section{Results}
\label{sec:results}

All numbers come from the full-study run records, scored by the frozen parser and placeholder rules except where marked post hoc.
Unless stated, they are on documents with $n < k$ at $k = 3$ and 5; 95\% group-bootstrap intervals are in the tables of contrasts.

\subsection{A shown minimum, per model}
\label{sec:main}

\begin{table*}[t]
  \centering
  \small
  \begin{adjustbox}{max width=\linewidth}\begin{tabular}{lrrrrr}
  \toprule
  Model & C1 & C2 & C3 & C4 & C5 \\
  \midrule
  Qwen3.5-0.8B & $-0.38$ [$-0.84$, $+0.07$] & $-0.59$ [$-1.06$, $-0.13$] & $-0.32$ [$-0.73$, $+0.08$] & $-0.63$ [$-0.77$, $-0.49$] & $-0.31$ [$-0.47$, $-0.16$] \\
  Qwen3.5-2B & $+0.12$ [$-1.39$, $+1.56$] & $+0.60$ [$-0.51$, $+1.72$] & $+2.13$ [$+1.22$, $+3.07$] & $-0.44$ [$-0.74$, $-0.13$] & $-0.22$ [$-0.32$, $-0.11$] \\
  Qwen3.5-9B & $+0.51$ [$+0.39$, $+0.65$] & $+0.62$ [$+0.50$, $+0.74$] & $+0.80$ [$+0.69$, $+0.92$] & $+0.11$ [$+0.06$, $+0.15$] & $-0.64$ [$-0.81$, $-0.53$] \\
  Qwen3.5-27B & $+1.20$ [$+1.07$, $+1.33$] & $+1.24$ [$+1.11$, $+1.37$] & $+1.29$ [$+1.18$, $+1.41$] & $+0.37$ [$+0.32$, $+0.41$] & $-0.57$ [$-0.61$, $-0.53$] \\
  Gemma 4 31B & $+1.13$ [$+0.99$, $+1.27$] & $+1.68$ [$+1.55$, $+1.81$] & $+1.70$ [$+1.57$, $+1.83$] & $+0.93$ [$+0.87$, $+0.99$] & $-0.66$ [$-0.70$, $-0.62$] \\
  Olmo 3.1 32B & $+0.47$ [$+0.29$, $+0.64$] & $+0.26$ [$+0.10$, $+0.42$] & $+1.20$ [$+1.00$, $+1.42$] & $+0.02$ [$-0.05$, $+0.08$] & $-0.64$ [$-0.73$, $-0.56$] \\
  Ministral 3 14B & $+1.03$ [$+0.87$, $+1.19$] & $+0.30$ [$+0.12$, $+0.45$] & $+1.13$ [$+0.99$, $+1.27$] & $+0.19$ [$+0.14$, $+0.24$] & $-0.94$ [$-1.00$, $-0.88$] \\
  gpt-oss-20b & $+0.83$ [$+0.65$, $+1.01$] & $+0.87$ [$+0.72$, $+1.01$] & $+0.98$ [$+0.86$, $+1.10$] & $+0.28$ [$+0.23$, $+0.33$] & $-0.32$ [$-0.38$, $-0.27$] \\
  \midrule
  Pooled & $+0.61$ [$+0.42$, $+0.80$] & $+0.62$ [$+0.46$, $+0.78$] & $+1.11$ [$+0.97$, $+1.25$] & $+0.10$ [$+0.06$, $+0.15$] & $-0.54$ [$-0.58$, $-0.50$] \\
  \bottomrule
\end{tabular}
\end{adjustbox}
  \caption{Confirmatory contrasts per GPU model and pooled (8 models, Holm $p < 0.0001$ for each pooled contrast). C1: \cnd{min\_k\_shown} at $k = 5$ minus $k = 3$ ($n \leq 2$). C2: \cnd{min\_k\_shown} minus \cnd{none}, the main measure. C3: \cnd{min\_k\_shown} minus \cnd{max\_k\_shown}. C4: \cnd{prose\_exactly\_k} minus \cnd{prose\_up\_to\_k}, five wordings averaged. C1 to C4 in substantive entries per output; C5: \cnd{min\_k\_escape} minus \cnd{min\_k\_enforced}, excess entries per forced slot. 95\% intervals from 10,000 bootstrap draws over document groups.}
  \label{tab:permodel}
\end{table*}

A minimum shown in the schema lengthens lists for six of the eight GPU models, by 0.26 (Olmo 3.1 32B) to 1.68 entries (Gemma 4 31B), each with an interval above 0 (Table~\ref{tab:permodel}, column C2).
Pooled over the eight, it adds 0.62 entries [0.46, 0.78] and raises over-count by 20.9 points [19.5, 22.3]: on the same documents, about one more list in five runs past the true count.
The estimate holds without the pilot documents, under the mixed model, with the five families weighted equally and with any one family left out (0.47 to 0.78; Appendices~\ref{app:generality} and~\ref{app:perset}).

The two smallest models are the exceptions.
Qwen3.5-2B lengthens its lists ($+0.60$), but its interval spans 0.
Qwen3.5-0.8B often writes no readable list: 36.6\% of its outputs are unreadable, mostly an answer written inside a copy of the schema, which the frozen rules count as 0 entries, so its estimate turns negative ($-0.59$; Appendix~\ref{app:outcomes}).
Read with a lenient parser, it over-counts on about half its outputs both with no count and under a shown minimum (post hoc; Appendix~\ref{app:posthoc}).
Because unreadable outputs count as empty lists, this model pulls the pooled numbers down: without it, the main contrast is $+0.80$ entries [0.63, 0.96] and over-count rises from 35.0\% to 59.8\% (post hoc).
The registered pooled estimate is therefore the conservative one.

The added entries are not true items that the no-count list missed: the share of true items left out (documents with $n \geq 1$) is 18.2\%\pre{} under a shown minimum and 17.0\%\pre{} with no count (Table~\ref{tab:outcomes}).
Over all documents, an enforced minimum misses the fewest true items (recall 0.91\pre{}, against 0.83\pre{} with no count) because it adds the most entries, and prose ``up to $k$'' has the best entry F1 (0.60\pre{}, against 0.53\pre{}; post hoc, Table~\ref{tab:prf}).

The other confirmatory contrasts agree (Table~\ref{tab:permodel}).
Raising the shown minimum from 3 to 5 on the same document adds 0.61 entries (C1), about 0.3 per added slot; over-count barely moves, since a list already above $n$ at $k = 3$ cannot over-count twice, which is why the registered outcome is length.
A shown minimum gives 1.11 entries more than a shown maximum (C3), though this sums two opposite effects ($+0.62$ and $-0.49$ against no count).
In prose, ``exactly $k$'' gives 0.10 entries and 8.2 over-count points more than ``up to $k$'' (C4).

A count in prose pulls lists toward $k$ from both sides.
In every wording, both prose conditions shorten lists against no count ($-0.43$ to $-1.02$ entries; Table~\ref{tab:wordings}), mostly by cutting long lists, yet the share that over-counts rises from 34.5\% to 42.7\% for ``up to $k$'' and 50.9\% for ``exactly $k$'' (Table~\ref{tab:outcomes}).
A mean length change can therefore hide padding, which is why over-count is reported beside it.
The prose conditions also replace the ``List all'' instruction, so part of their difference from no count may come from the wording rather than the count.

The effect holds with the annotators' original counts, on documents whose list the cleanup kept or changed, and on $n = 0$ as on $n \geq 1$ documents (Appendix~\ref{app:key}).
By set, contrast 2 runs from $+0.07$ (Wikipedia people) to $+1.54$ entries (job-ad skills), and over-count rises in all nine sets, by 6.5 to 31.2 points (Appendix~\ref{app:perset}).

\subsection{Registered controls and the empty-document test}
\label{sec:named}

\begin{table}[t]
  \centering
  \small
  \setlength{\tabcolsep}{3pt}
  \begin{adjustbox}{max width=\linewidth}
  \begin{tabular}{@{}llr@{}}
    \toprule
    Test & Unit & Estimate [95\% CI] \\
    \midrule
      \cnd{fewer\_ok} vs \cnd{none} & entries & $-$0.77 [$-$0.93, $-$0.60] \\
  \cnd{fewer\_ok} vs \cnd{min\_k\_shown} & entries & $-$1.39 [$-$1.53, $-$1.25] \\
  \cnd{fewer\_ok} vs \cnd{none} & points & $-$3.5 [$-$4.7, $-$2.4] \\
  \cnd{max\_k\_shown} vs \cnd{none} & entries & $-$0.49 [$-$0.61, $-$0.37] \\
  $n \geq k$: \cnd{min\_k\_shown} vs \cnd{none} & entries & $-$0.15 [$-$0.30, $-$0.02] \\
  $n \geq k$: true items left out$^{\dagger}$ & points & +0.2 [$-$0.3, +0.8] \\
  $n < k$ minus $n \geq k$, $k = 3, 5$ & entries & +0.93 [+0.63, +1.24] \\
 
    \bottomrule
  \end{tabular}
  \end{adjustbox}

  \vspace{6pt}
  \begin{adjustbox}{max width=\linewidth}
  \begin{tabular}{@{}lrr@{}}
    \toprule
    Empty documents, $k = 1$ & $\geq 1$ entry \% & vs \cnd{none} (pts) \\
    \midrule
      \cnd{none} & 34.9 & -- \\
  \cnd{prose\_up\_to\_k} & 43.7 & +8.8 [+7.3, +10.3] \\
  \cnd{prose\_exactly\_k} & 46.5 & +11.6 [+9.8, +13.4] \\
  \cnd{max\_k\_shown} & 33.9 & $-$1.0 [$-$2.4, +0.4] \\
  \cnd{min\_k\_shown} & 42.3 & +7.4 [+5.6, +9.2] \\
  \cnd{min\_k\_shown\_fewer\_ok} & 31.8 & $-$3.1 [$-$5.0, $-$1.2] \\
  \cnd{min\_k\_enforced} & 82.4 & +47.5 [+44.5, +50.5] \\
  \cnd{min\_k\_escape} & 30.0 & $-$4.9 [$-$6.9, $-$2.9] \\
 
    \bottomrule
  \end{tabular}
  \end{adjustbox}
  \caption{Registered controls and the empty-document test, eight GPU models pooled. Top: documents with $n < k$, $k = 3$ and 5, except the two $n \geq k$ rows ($k = 1$, 3 and 5); $n < k$ minus $n \geq k$ is the change on $n < k$ documents minus the change on $n \geq k$ documents. Bottom: the 105 documents with no true item at $k = 1$, share of outputs with at least one substantive entry. \pre{} rests on the matcher. Per model: Appendix~\ref{app:named}.}
  \label{tab:named}
\end{table}

\paragraph{Fewer is fine.}
Adding ``If the document contains fewer than $k$, list all available ones.'' reverses the effect: lists end 0.77 entries shorter than with no count, and over-count falls 3.5 points below the no-count level (Table~\ref{tab:named}).
Seven of the eight models shorten their lists against the plain shown minimum, though Qwen3.5-27B still lists 0.46 entries more than with no count (Appendix~\ref{app:named}).
The line has a cost: the share of true items left out rises from 17.0\%\pre{} to 24.3\%\pre{} (Table~\ref{tab:outcomes}).

\paragraph{Enough items.}
On documents that already hold at least $k$ true items, a shown minimum does not lengthen lists ($-0.15$ entries [$-0.30$, $-0.02$]) and leaves true items left out unchanged.
On $n < k$ documents its effect exceeds this by 0.93 entries: the effect comes from filling the forced slots.

\paragraph{Empty documents at $k = 1$.}
On the 105 documents with no true item, even \texttt{minItems: 1}, which asks only for a non-empty list, raises the share of outputs with at least one substantive entry from 34.9\% with no count by 7.4 points when shown, and by 47.5 points, to 82.4\%, when enforced.
The ``fewer is fine'' line and the N/A escape both bring it below the no-count level.

\paragraph{maxItems.}
A shown maximum is not neutral: it shortens lists against no count ($-0.49$ entries [$-0.61$, $-0.37$]), so it fails the protocol's test for a harmless maximum.
It barely changes over-count (35.7\% against 34.5\%) but leaves out more true items (20.9\%\pre{} against 17.0\%\pre{}).

\subsection{What the lists hold}
\label{sec:what}

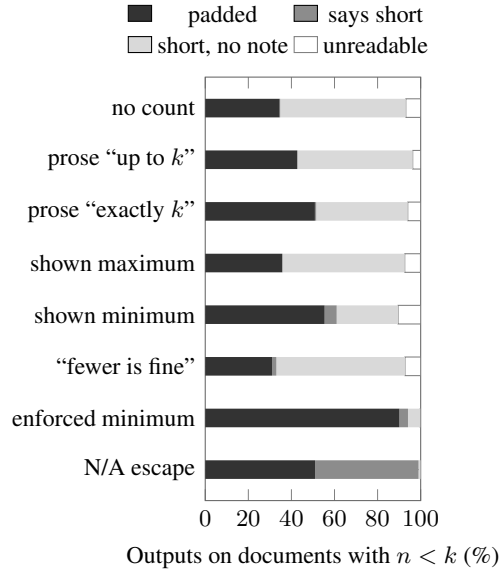
\begin{figure}[t]
  \centering
  \begin{tikzpicture}
    \begin{axis}[xbar stacked, scale only axis, width=0.37\columnwidth, height=5.6cm, bar width=7pt,
                 xmin=0, xmax=100, xlabel={Outputs on documents with $n < k$ (\%)},
                 ytick={1,...,8}, ymin=0.4, ymax=8.6,
                 yticklabels={\cnd{min\_k\_escape}, \cnd{min\_k\_enforced}, \cnd{min\_k\_shown\_fewer\_ok}, \cnd{min\_k\_shown}, \cnd{max\_k\_shown}, \cnd{prose\_exactly\_k}, \cnd{prose\_up\_to\_k}, \cnd{none}},
                 y tick label style={font=\footnotesize}, x tick label style={font=\footnotesize},
                 label style={font=\footnotesize}, axis line style={black!60},
                 legend style={font=\footnotesize, draw=none, at={(0.35,1.02)}, anchor=south, legend columns=2}]
      \addplot[fill=black!80, draw=none] coordinates {(51.0,1) (90.1,2) (31.0,3) (55.4,4) (35.7,5) (50.9,6) (42.7,7) (34.5,8)};
      \addplot[fill=black!45, draw=none] coordinates {(48.0,1) (4.0,2) (2.0,3) (5.6,4) (0.2,5) (0.6,6) (0.1,7) (0.3,8)};
      \addplot[fill=black!15, draw=none] coordinates {(0.9,1) (5.7,2) (59.7,3) (28.5,4) (56.7,5) (42.5,6) (53.5,7) (58.3,8)};
      \addplot[fill=white, draw=black!50] coordinates {(0.1,1) (0.2,2) (7.3,3) (10.5,4) (7.3,5) (6.0,6) (3.7,7) (6.9,8)};
      \legend{padded, says short, {short, no note}, unreadable}
    \end{axis}
  \end{tikzpicture}
  \caption{How outputs on documents with $n < k$ end, by condition, eight GPU models pooled, $k = 3$ and 5. Padded: more substantive entries than $n$, even with a note. All columns: Table~\ref{tab:outcomes}.}
  \label{fig:sorting}
\end{figure}

Over-count is not zero without a count: 34.5\% of no-count outputs on $n < k$ documents already list more than $n$ substantive entries, mostly grounded padding (64.6\%\pre{}; Table~\ref{tab:fillfull}).
This is ordinary extraction error, and each paired contrast measures what a condition adds to it on the same document and model.
On the documents a model gets exactly right with no count, a shown minimum makes 35.2\% [33.0, 37.5] of its lists pad, against 5.0\% under a shown maximum (post hoc; Table~\ref{tab:cleanbase}).
A shown minimum moves lists toward $k$: the share with exactly $k$ entries rises from 5.6\% to 35.5\%, and 55.4\% of outputs pad (Figure~\ref{fig:sorting}).
Enforced, the minimum leaves 0.6\% of lists short of $k$ and 90.1\% padded.
That much follows from the decoder, which all but rules out a shorter list; what the enforced condition shows is what fills the slots.

Most extra entries are grounded padding: 62.9\%\pre{} under a shown minimum and 79.6\%\pre{} under an enforced one, as in Figure~\ref{fig:example}.
Most of the rest are invented: 29.5\%\pre{} under a shown minimum, about the share with no count (29.8\%\pre{}), so the minimum adds both kinds in the proportions they already have; corrected with the authors' labels, the invented share is 23.4\% (post hoc; Appendix~\ref{app:posthoc}).
That makes the padding hard to catch: a source check removes invented entries, but 46.6\%\pre{} of all extra entries under a shown minimum pass it (post hoc).

\paragraph{The N/A escape (C5).}
Offering ``N/A'' under an enforced minimum cuts the excess entries per forced slot by 0.54 (rate ratio 0.57 in the Poisson mixed model), and every GPU model's own estimate is below 0 (Table~\ref{tab:permodel}).
It is not a full fix: 51.0\% of outputs still pad, and the two smallest models barely use it (Qwen3.5-0.8B still pads 89.6\% of its outputs and Qwen3.5-2B 83.0\%; Appendix~\ref{app:outcomes}).

\subsection{Hosted models and serving stack}
\label{sec:hosted}

All nine hosted runs lengthen their lists under a shown minimum, by 0.28 (Ministral 3 14B on Bedrock) to 2.39 entries (DeepSeek V3.2; Figure~\ref{fig:headline}), and give more entries under a minimum than under a maximum (Table~\ref{tab:hosted}).
For four of them (both DeepSeek models, Kimi K3 and Qwen3-235B), the ``fewer is fine'' line does not undo the effect: they still list 0.30 to 0.53 entries more than with no count.
The serving stack does not drive the result: for the three models run both on the GPU and hosted, the main contrast differs by at most 0.04 entries and 1.1 over-count points, within the registered limit of 10 points.

\subsection{Exploratory: re-asking and thinking}
\label{sec:exploratory}

\paragraph{Re-asking.}
Libraries that validate a schema send a failing reply back with the validation error.
We did this once for every shown-minimum reply shorter than $k$, in five runs (four models, with Ministral 3 14B both on the GPU and on Bedrock).
Over-count among the re-asked outputs rose in every run: from 10.0\% to 69.0\% for Qwen3.5-9B and from 14.0\% to 78.3\% for Kimi K3 (Appendix~\ref{app:reask}).
The error restates the minimum, so complying with it is expected.
The point is practical: after the default retry, 44.0 to 81.6\% of these short lists, truthful ones included, are padded.
A re-ask that adds ``fewer is fine'' was not tested.

\paragraph{Thinking.}
Thinking does not help (Appendix~\ref{app:thinking}, Table~\ref{tab:thinking}).
For Qwen3.5-9B it makes padding worse: in each of three request seeds, contrast 2 is 1.1 to 1.2 entries larger with thinking on, and over-count under a shown minimum rises from 43--47\% to 81--83\%.
Kimi K3 moves the same way, less ($+0.58$ entries on the pilot documents; its pair also differs in temperature).
Gemma 4 31B shows no length effect ($-0.11$ [$-0.23$, 0.01]), though its over-count under a shown minimum rises from 61.8\% to 66.1\%.
The thinking often names the shortfall, in 88--89\% of Qwen3.5-9B's padded outputs by a phrase search, and the model pads anyway.

\section{Discussion}
\label{sec:discussion}

\paragraph{Is this just instruction following?}
In part it is: the model follows the schema over ``List all'', and Section~\ref{sec:scoring} gives our reasons for counting that as the failure.
The controls show that the minimum does this, not the schema or any number in it.
Under a shown maximum 35.7\% of outputs over-count, against 34.5\% with no count and 55.4\% under a shown minimum.
A minimum adds no entries when the document already holds enough items.
Told that a shorter list is fine, seven of eight GPU models shorten their lists against the plain minimum, so most can stop.
What goes wrong is that the model meets the count with whatever text is at hand instead of saying the list is short.

\paragraph{Why it is easy to miss.}
Most of the padding is taken from the document: it parses, meets the schema, and about half of it passes a source check (Section~\ref{sec:what}), so the error shows in full only against an answer list.
This matches what \citet{han2023empirical} and \citet{qian2026unreliable} find for extraction without a minimum count, and differs from PhantomFill, where the inputs hold nothing to copy \citep{usman2026phantomfill}.

\begin{table}[t]
  \centering
  \small
  \fbox{\begin{minipage}{0.92\columnwidth}
  \textbf{For schema authors}:
  \begin{itemize}
    \setlength\itemsep{0pt}
    \item Leave \texttt{minItems} off lists whose length the document decides (over-count 34.5\% $\to$ 55.4\% shown, 90.1\% enforced).
    \item If recall matters most, an enforced minimum misses the fewest true items but pads most; prose ``up to $k$'' balances the two best (F1 0.60\pre{} vs 0.53\pre{}).
    \item Do not rely on a source check: about half of the padding passes it.
    \item If a minimum must stay, add ``fewer is fine'' and test it: some models still lengthen their lists, and recall falls.
    \item An ``N/A'' option under enforcement helps but is not a fix (90.1\% $\to$ 51.0\% padded).
    \item Do not re-ask with the bare validation error (Qwen3.5-9B: 10.0\% $\to$ 69.0\% over-count).
    \item \texttt{maxItems} barely changes over-count but shortens lists and leaves out more true items.
  \end{itemize}
  \end{minipage}}
  \caption{Guidance from the open-weight models tested.}
  \label{tab:guidance}
\end{table}

\section{Conclusion}
\label{sec:conclusion}

When a document holds fewer items than a schema's minimum, most of the open-weight models tested treat the count as a target and pad their lists: a shown minimum raises the share of over-counting lists from 34.5\% to 55.4\%, and an ``N/A'' option under an enforced minimum still leaves half of them padded.
The added entries are mostly real text taken from the document, so about half of them pass a source check.
A ``fewer is fine'' line undoes the effect on average at some cost in recall, while re-asking with the validation error makes padding far more common, and thinking does not help.

\section*{Limitations}

\paragraph{Scope.}
Every list item is one string (a quote, a name or, for meetings, a summary sentence); lists of records with several fields, where padding would force invented field values, are not tested.
One grammar engine (XGrammar) is used; a second engine gave different outputs in a pre-study test.
The schema is always pasted in the prompt, in LangChain's format, and $k$ takes the values 1, 3 and 5.
No closed model is run, and closed APIs may pad less (Section~\ref{sec:models}); the six conditions with a free decoder run on any API, so they are the natural next test.

\paragraph{Answer lists and matcher.}
The answer lists fix one reading for each ambiguous case, so the blind check measures agreement with the written rules; job-ad skills is the least certain set.
For this reason the main measure needs only the true count, and the confirmatory contrasts are also given with the annotators' original counts.
Numbers marked \pre{} rest on the item matcher, which does not assign the labels ``true item the list missed'' and ``text from beyond the excerpt''; its invented shares are its least firm numbers.
Thinking traces are read only by a phrase search, not by hand.
The share of outputs that note a shortfall outside the list is a lower bound, because the frozen note rules miss some wordings.

\paragraph{Setup.}
Decoding is greedy outside the thinking test, but repeated prompts did not always give the same reply: the over-count verdict changed for at most 3.0\% of GPU prompts and 8.0\% of Bedrock prompts sent twice (Appendix~\ref{app:repeat}).
Two chat templates add a system prompt of their own (Olmo 3.1 a default one, gpt-oss one with the current date).
Hosted models are a reference, with no pinned revision or serving stack.
The novel texts lost their non-ASCII characters; the CORD receipts are Indonesian and the other sets English.
The documents fill fixed cells of up to 15 per set and true count, and documents with no single right count were dropped, so how often a real pipeline meets $n < k$ is not measured here.
In public schemas a minimum on a list of strings or objects is almost always 1 (Appendix~\ref{app:posthoc}), so the $k = 1$ test matches the common case and $k = 3$ and 5 a rarer one.

\section*{Ethics statement}

All nine sets come from public datasets, used under their licences (Table~\ref{tab:licences}).
The reviews set's annotations are Apache-2.0, but its review text states no licence, so it is not redistributed: those documents are rebuilt locally from the original source, and the release carries only their identifiers and text hashes.
The Enron emails are real people's mail, public since 2003; the clinical case reports are published in PubMed Central.
MailEx, Few-NERD and the Stack Overflow (TECH) job ads of SkillSpan are share-alike, so a released benchmark that contains their text is licensed CC BY-SA 4.0.
The hosted runs sent the study prompts, review documents included, to Together AI and Amazon Bedrock for inference.
All human labelling and checking was done by the two authors; no crowd workers or other participants were involved.

\section*{AI-use statement}

We used an AI coding assistant (Kiro, running Claude Opus 5.5) for implementation and drafting.
The authors conceived the study, made every design decision, and did every human judgement in it: we set the counting rules and settled the cases they left open, did the blind checks, labelled the 400 matcher-validation entries independently, and checked or recounted 135 scored outputs by hand.
The assistant wrote and tested the code, ran and logged the experiments, checked citations against the cited papers, and drafted text that we revised.
It also made a first pass applying our counting rules to the 1,446 candidate answer lists; author B then checked every resulting decision by hand.
No AI output was used as a human label, and the item matcher is evaluated only against our labels.
No Claude model is among the models tested.
We take full responsibility for the content.

\section*{Acknowledgements}

We thank the creators and annotators of the datasets this study builds on (Appendix~\ref{app:data}), and the maintainers of vLLM, XGrammar and the other open-source tools it runs on.
The Amazon Bedrock runs used AWS promotional credits.

\bibliography{refs}

\begin{thebibliography}{23}
\providecommand{\natexlab}[1]{#1}

\bibitem[{Brach et~al.(2026)Brach, Zuppichini, Vinciguerra, and
  Padoan}]{brach2026scrapegraphai}
William Brach, Francesco Zuppichini, Marco Vinciguerra, and Lorenzo Padoan.
  2026.
\newblock \href {https://arxiv.org/abs/2602.15189} {{ScrapeGraphAI-100k:
  Dataset for Schema-Constrained LLM Generation}}.
\newblock \emph{Preprint}, arXiv:2602.15189.

\bibitem[{Carletta et~al.(2006)Carletta, Ashby, Bourban, Flynn, Guillemot,
  Hain, Kadlec, Karaiskos, Kraaij, Kronenthal, Lathoud, Lincoln, Lisowska,
  McCowan, Post, Reidsma, and Wellner}]{carletta2005ami}
Jean Carletta, Simone Ashby, Sebastien Bourban, Mike Flynn, Mael Guillemot,
  Thomas Hain, Jaroslav Kadlec, Vasilis Karaiskos, Wessel Kraaij, Melissa
  Kronenthal, Guillaume Lathoud, Mike Lincoln, Agnes Lisowska, Iain McCowan,
  Wilfried Post, Dennis Reidsma, and Pierre Wellner. 2006.
\newblock \href {https://doi.org/10.1007/11677482_3} {{The AMI Meeting Corpus:
  A Pre-announcement}}.
\newblock In \emph{{Machine Learning for Multimodal Interaction}}, volume 3869
  of \emph{{Lecture Notes in Computer Science}}, pages 28--39, Berlin,
  Heidelberg. Springer.
\newblock MLMI 2005.

\bibitem[{Caufield et~al.(2019)Caufield, Zhou, Bai, Liem, Garlid, Chang, Sun,
  Ping, and Wang}]{caufield2019maccrobat}
J.~Harry Caufield, Yichao Zhou, Yunsheng Bai, David~A. Liem, Anders~O. Garlid,
  Kai-Wei Chang, Yizhou Sun, Peipei Ping, and Wei Wang. 2019.
\newblock \href {https://doi.org/10.1101/19009118} {{A Comprehensive Typing
  System for Information Extraction from Clinical Narratives}}.
\newblock \emph{{medRxiv}}.

\bibitem[{Chebolu et~al.(2024)Chebolu, Dernoncourt, Lipka, and Solorio}]{roast}
Siva Uday~Sampreeth Chebolu, Franck Dernoncourt, Nedim Lipka, and Thamar
  Solorio. 2024.
\newblock \href {https://doi.org/10.48550/arXiv.2405.20274} {{ROAST:
  Review-level Opinion Aspect Sentiment Target Joint Detection for ABSA}}.
\newblock \emph{{arXiv preprint arXiv:2405.20274}}.

\bibitem[{Ding et~al.(2021)Ding, Xu, Chen, Wang, Han, Xie, Zheng, and
  Liu}]{ding2021fewnerd}
Ning Ding, Guangwei Xu, Yulin Chen, Xiaobin Wang, Xu~Han, Pengjun Xie, Haitao
  Zheng, and Zhiyuan Liu. 2021.
\newblock \href {https://doi.org/10.18653/v1/2021.acl-long.248} {{Few-NERD: A
  Few-shot Named Entity Recognition Dataset}}.
\newblock In \emph{{Proceedings of the 59th Annual Meeting of the Association
  for Computational Linguistics and the 11th International Joint Conference on
  Natural Language Processing (Volume 1: Long Papers)}}, pages 3198--3213,
  Online. Association for Computational Linguistics.

\bibitem[{Dong et~al.(2025)Dong, Ruan, Cai, Lai, Xu, Zhao, and
  Chen}]{dong2025xgrammar}
Yixin Dong, Charlie~F. Ruan, Yaxing Cai, Ruihang Lai, Ziyi Xu, Yilong Zhao, and
  Tianqi Chen. 2025.
\newblock \href {https://arxiv.org/abs/2411.15100} {{XGrammar: Flexible and
  Efficient Structured Generation Engine for Large Language Models}}.
\newblock In \emph{Proceedings of Machine Learning and Systems}, volume~7.

\bibitem[{Geng et~al.(2025)Geng, Cooper, Moskal, Jenkins, Berman, Ranchin,
  West, Horvitz, and Nori}]{geng2025jsonschemabench}
Saibo Geng, Hudson Cooper, Micha{\l} Moskal, Samuel Jenkins, Julian Berman,
  Nathan Ranchin, Robert West, Eric Horvitz, and Harsha Nori. 2025.
\newblock \href {https://arxiv.org/abs/2501.10868} {{JSONSchemaBench: A
  Rigorous Benchmark of Structured Outputs for Language Models}}.
\newblock \emph{Preprint}, arXiv:2501.10868.

\bibitem[{Han et~al.(2024)Han, Yang, Peng, Tiwari, Wan, Liu, and
  Wang}]{han2023empirical}
Ridong Han, Chaohao Yang, Tao Peng, Prayag Tiwari, Xiang Wan, Lu~Liu, and
  Benyou Wang. 2024.
\newblock \href {https://arxiv.org/abs/2305.14450} {{An Empirical Study on
  Information Extraction using Large Language Models}}.
\newblock \emph{Preprint}, arXiv:2305.14450.
\newblock Version 2; version 1 (2023) was titled ``Is Information Extraction
  Solved by ChatGPT? An Analysis of Performance, Evaluation Criteria,
  Robustness and Errors''.

\bibitem[{Ivgi et~al.(2024)Ivgi, Yoran, Berant, and Geva}]{ivgi2025loops}
Maor Ivgi, Ori Yoran, Jonathan Berant, and Mor Geva. 2024.
\newblock \href {https://arxiv.org/abs/2407.06071} {{From Loops to Oops:
  Fallback Behaviors of Language Models Under Uncertainty}}.
\newblock In \emph{NeurIPS 2024 Workshop on Attributing Model Behavior at Scale
  (ATTRIB)}.

\bibitem[{Kirichenko et~al.(2025)Kirichenko, Ibrahim, Chaudhuri, and
  Bell}]{kirichenko2025abstentionbench}
Polina Kirichenko, Mark Ibrahim, Kamalika Chaudhuri, and Samuel~J. Bell. 2025.
\newblock \href {https://arxiv.org/abs/2506.09038} {{AbstentionBench: Reasoning
  LLMs Fail on Unanswerable Questions}}.
\newblock \emph{Preprint}, arXiv:2506.09038.

\bibitem[{Kwon et~al.(2023)Kwon, Li, Zhuang, Sheng, Zheng, Yu, Gonzalez, Zhang,
  and Stoica}]{kwon2023vllm}
Woosuk Kwon, Zhuohan Li, Siyuan Zhuang, Ying Sheng, Lianmin Zheng, Cody~Hao Yu,
  Joseph~E. Gonzalez, Hao Zhang, and Ion Stoica. 2023.
\newblock \href {https://doi.org/10.1145/3600006.3613165} {{Efficient Memory
  Management for Large Language Model Serving with PagedAttention}}.
\newblock In \emph{Proceedings of the 29th Symposium on Operating Systems
  Principles (SOSP '23)}, pages 611--626. ACM.

\bibitem[{Lee et~al.(2026)Lee, D'Antoni, and Berg-Kirkpatrick}]{lee2026format}
Ivan~Yee Lee, Loris D'Antoni, and Taylor Berg-Kirkpatrick. 2026.
\newblock \href {https://arxiv.org/abs/2604.03616} {{The Format Tax}}.
\newblock \emph{Preprint}, arXiv:2604.03616.

\bibitem[{Park et~al.(2024)Park, Wang, Berg-Kirkpatrick, Polikarpova, and
  D'Antoni}]{park2026grammar}
Kanghee Park, Jiayu Wang, Taylor Berg-Kirkpatrick, Nadia Polikarpova, and Loris
  D'Antoni. 2024.
\newblock \href {https://arxiv.org/abs/2405.21047} {{Grammar-Aligned
  Decoding}}.
\newblock In \emph{Advances in Neural Information Processing Systems},
  volume~37.

\bibitem[{Park et~al.(2019)Park, Shin, Lee, Lee, Surh, Seo, and
  Lee}]{park2019cord}
Seunghyun Park, Seung Shin, Bado Lee, Junyeop Lee, Jaeheung Surh, Minjoon Seo,
  and Hwalsuk Lee. 2019.
\newblock \href {https://openreview.net/forum?id=SJl3z659UH} {{CORD: A
  Consolidated Receipt Dataset for Post-OCR Parsing}}.
\newblock In \emph{{Workshop on Document Intelligence at NeurIPS 2019}}.

\bibitem[{Qian(2026)}]{qian2026unreliable}
Jiahong Qian. 2026.
\newblock \href {https://arxiv.org/abs/2610.20417} {{From Unreliable to Usable:
  Decomposing Failure Modes and Confidence Signals in LLM Entity Extraction --
  Evidence from Cyber Threat Intelligence}}.
\newblock \emph{Preprint}, arXiv:2610.20417.

\bibitem[{Srivastava et~al.(2023)Srivastava, Singh, Matsumoto, Raz, Costa,
  Poore, and Yao}]{srivastava2023mailex}
Saurabh Srivastava, Gaurav Singh, Shou Matsumoto, Ali Raz, Paulo Costa, Joshua
  Poore, and Ziyu Yao. 2023.
\newblock \href {https://doi.org/10.18653/v1/2023.emnlp-main.801} {{MailEx:
  Email Event and Argument Extraction}}.
\newblock In \emph{{Proceedings of the 2023 Conference on Empirical Methods in
  Natural Language Processing}}, pages 12964--12987, Singapore. Association for
  Computational Linguistics.

\bibitem[{Tam et~al.(2024)Tam, Wu, Tsai, Lin, yi~Lee, and Chen}]{tam2024speak}
Zhi~Rui Tam, Cheng-Kuang Wu, Yi-Lin Tsai, Chieh-Yen Lin, Hung yi~Lee, and
  Yun-Nung Chen. 2024.
\newblock \href {https://arxiv.org/abs/2408.02442} {{Let Me Speak Freely? A
  Study on the Impact of Format Restrictions on Large Language Model
  Performance}}.
\newblock In \emph{Proceedings of the 2024 Conference on Empirical Methods in
  Natural Language Processing: Industry Track}.

\bibitem[{Usman(2026)}]{usman2026phantomfill}
Rana~Muhammad Usman. 2026.
\newblock \href {https://arxiv.org/abs/2607.20492} {{PhantomFill: When the Form
  Demands an Answer, Language Models Invent One}}.
\newblock \emph{Preprint}, arXiv:2607.20492.

\bibitem[{Vishnubhotla et~al.(2022)Vishnubhotla, Hammond, and
  Hirst}]{vishnubhotla2022pdnc}
Krishnapriya Vishnubhotla, Adam Hammond, and Graeme Hirst. 2022.
\newblock \href {https://aclanthology.org/2022.lrec-1.628/} {{The Project
  Dialogism Novel Corpus: A Dataset for Quotation Attribution in Literary
  Texts}}.
\newblock In \emph{{Proceedings of the Thirteenth Language Resources and
  Evaluation Conference}}, pages 5838--5848, Marseille, France. European
  Language Resources Association.

\bibitem[{Wen et~al.(2025)Wen, Yao, Feng, Xu, Tsvetkov, Howe, and
  Wang}]{wen2025know}
Bingbing Wen, Jihan Yao, Shangbin Feng, Chenjun Xu, Yulia Tsvetkov, Bill Howe,
  and Lucy~Lu Wang. 2025.
\newblock \href {https://doi.org/10.1162/tacl_a_00754} {{Know Your Limits: A
  Survey of Abstention in Large Language Models}}.
\newblock \emph{Transactions of the Association for Computational Linguistics},
  13:529--556.

\bibitem[{Yang(2026)}]{yang2026prompt}
Dekun Yang. 2026.
\newblock \href {https://arxiv.org/abs/2607.01240} {{Prompt Framing Distorts
  Count-Based Evaluation of LLM Error Detection: Evidence from Numeric
  Anchoring}}.
\newblock \emph{Preprint}, arXiv:2607.01240.

\bibitem[{Zhang et~al.(2022)Zhang, Jensen, Sonniks, and
  Plank}]{zhang2022skillspan}
Mike Zhang, Kristian Jensen, Sif Sonniks, and Barbara Plank. 2022.
\newblock \href {https://doi.org/10.18653/v1/2022.naacl-main.366} {{SkillSpan:
  Hard and Soft Skill Extraction from English Job Postings}}.
\newblock In \emph{{Proceedings of the 2022 Conference of the North American
  Chapter of the Association for Computational Linguistics: Human Language
  Technologies}}, pages 4962--4984, Seattle, United States. Association for
  Computational Linguistics.

\bibitem[{Zhou et~al.(2023)Zhou, Lu, Mishra, Brahma, Basu, Luan, Zhou, and
  Hou}]{zhou2023instruction}
Jeffrey Zhou, Tianjian Lu, Swaroop Mishra, Siddhartha Brahma, Sujoy Basu,
  Yi~Luan, Denny Zhou, and Le~Hou. 2023.
\newblock \href {https://arxiv.org/abs/2311.07911} {{Instruction-Following
  Evaluation for Large Language Models}}.
\newblock \emph{Preprint}, arXiv:2311.07911.

\end{thebibliography}

\appendix

\section{Deviations from the registered plan}
\label{app:deviations}

Table~\ref{tab:deviations} lists every place where the analysis could not follow \texttt{preregistration/protocol-stage2.md} (tag \texttt{prereg-v2.1}) as written.
The changes made after the pilot and before the registration are part of the registered plan (its section 10) and are not repeated here.

\begin{table*}[t]
  \centering
  \small
  \begin{adjustbox}{max width=\linewidth}
  \begin{tabular}{@{}p{0.30\textwidth}p{0.64\textwidth}@{}}
    \toprule
    Registered & What was done \\
    \midrule
    Matcher thresholds frozen and tagged \texttt{scorer-v1} before study outputs are scored (5.2) & The matcher was frozen with the registered plan (tag \texttt{prereg-v2.1}) instead of under a separate tag, and scores the study unchanged. \\
    Five matcher labels (5.2, 5.3 b) & The matcher does not assign ``true item the list missed'' or ``text from beyond the excerpt''; those entries land in ``wrong kind'' or ``invented''. Their places in the 400-entry draw go in equal parts to the other three labels (134, 133, 133), as 5.3 b allows for a label with too few entries. \\
    Scoring check on 40 pilot outputs before the full run (5.3 a) & Drawn as registered and done during the full run rather than before it (65 replies: 40 random, 3 cut at the token limit, 22 unreadable). No scorer bug was found (Appendix~\ref{app:checks}). \\
    Escape outcome ``extra entries per padded slot'' (6.2, contrast 5) & Computed per output as excess / ($k - n$), the slots the grammar forces beyond the true items; this paper calls them forced slots. \\
    Poisson mixed model for contrast 5 (6.2) & A variational-Bayes fit with a random intercept per document and $\log(k - n)$ as a covariate, reported as a rate ratio next to the primary bootstrap test. \\
    Mixed models of the paired difference (6.2) & Covariates centred and factors sum-coded, so the intercept is the average paired difference; the intercept and its Wald interval are reported. \\
    ``Says it is short'' from notes outside the list (5.2) & The frozen note rules applied to each sentence outside the JSON; notes they do not cover are missed, so the share is a lower bound. \\
    $n = 0$ analysed apart (6.1) & The primary contrasts pool every $n < k$ cell, as the pilot's stop rules did; $n = 0$ and $n \geq 1$ are given separately (Appendix~\ref{app:key}). \\
    Same-weights pairs, ``10 points'' (3.3) & The rule was written for over-count; the pair table gives length and over-count and judges the rule on over-count points. \\
    400-entry frame: every substantive entry of every study output (5.3 b) & The frame leaves out the thinking-pair runs, which finished after the draw was fixed. \\
    RQ5, RQ7, RQ8 and the robustness arms (2, 3.2) & The teacher-forced and hidden-state analyses are left to future work; the robustness arms ran in the pilot. \\
    Every request answered (9) & 11 outputs are missing after repeated transport errors (Qwen3.5-0.8B 1, Qwen3.5-2B 3, gpt-oss-20b 6) or timeouts (Gemma 4 31B with thinking on, 1); they are left out of every pair. \\
    \bottomrule
  \end{tabular}
  \end{adjustbox}
  \caption{Deviations from the registered plan (protocol section in brackets).}
  \label{tab:deviations}
\end{table*}

\section{The authors' checks}
\label{app:checks}

The protocol registers three checks by the authors (section 5.3); their sheets were drawn with a fixed seed by committed scripts (\texttt{analysis/full\_study\_sheets.py}, \texttt{analysis/human/}).

\paragraph{(a) Scoring check.}
One author reads 65 pilot replies (40 drawn at random, the 3 outputs cut at the token limit and all 22 unreadable ones) against the scorer: parse rung, substantive entries, placeholders, repeats, over-count.
The author and the scorer agree fully on 61 of the 65.
The 4 differences all come from how the frozen scorer reads an output.
In one, the author counts as a repeat an entry that restates an earlier one with a three-word prefix; the scorer counts as repeats only entries with the same or nearly the same words (overlap threshold 0.85, here 0.67) and leaves the rest to the matcher.
In three, the reply is cut at the token limit inside an entry; the author counts the unfinished entry and the scorer keeps only entries that close before the cut, so in one of them the author sees an over-count the scorer does not.
No scorer bug was found, so nothing is rescored.

\paragraph{(b) Matcher labels.}
The matcher matches names after normalising case and punctuation, quotes by fuzzy substring search, and meeting action items by similarity to each answer item and its transcript lines.
Both authors label 400 entries alone, blind to the model, condition, $k$, the other entries and the matcher's label, then settle the differences.
They chose the same label for 370 of the 400 entries (92.5\%; Cohen's kappa 0.896, Gwet's AC1 0.912) and settled the other 30 together; the settled label is the reference.
Table~\ref{tab:matchercheck} gives the matcher against it.
The matcher gave no study entry the labels ``true item the list missed'' or ``text from beyond the excerpt'', so those strata are empty and their places went to the other three.
The invented share of substantive entries is 7.0\% as the matcher gives it and 7.3\% weighted by the authors' labels.
Five entries were settled as true items the list missed; they go into the answer-list overlay with the document's wording and the authors' reason (\texttt{analysis/gold\_overlay.jsonl}; the one review item's text is in the local restricted file), and with the overlay applied contrasts 1 to 4 are the same to two decimals and contrast 2 on over-count moves from $+20.9$ to $+21.1$ points.

\begin{table*}[t]
  \centering
  \small
  \begin{tabular}{@{}lrrrcc@{}}
    \toprule
    Matcher label & Frame entries & Drawn & Used & Precision [95\% CI] & Recall [95\% CI] \\
    \midrule
    True item or split & 637,844 & 134 & 134 & 0.98 [0.95, 1.00] & 0.94 [0.90, 0.97] \\
    Wrong kind & 408,859 & 133 & 133 & 0.85 [0.78, 0.91] & 0.96 [0.93, 0.99] \\
    Invented & 78,798 & 133 & 131 & 0.69 [0.56, 0.83] & 0.66 [0.53, 0.83] \\
    \bottomrule
  \end{tabular}
  \caption{The matcher against the authors' settled labels of 400 entries. Precision: share of the matcher's calls the authors confirm. Recall: share of the authors' labels the matcher finds, each stratum weighted back to its size in the frame. Intervals: 10,000 bootstrap draws over document groups. Used: entries whose settled label is not ``unsure''; the two left out are clinical entries saying no medication was given, which the authors settled as unsure.}
  \label{tab:matchercheck}
\end{table*}

\paragraph{(c) Hand recount.}
For 10 documents with $n < k$, each paired with one GPU model, one author counts the substantive entries of 7 outputs by hand (70 outputs) and computes length change, over-count, true items left out and the sorting.
The hand count agrees fully with the scorer on 60 of the 70.
In 7 (Qwen3.5-0.8B on one email), the model wrote its list inside a copy of the schema; the frozen parser counts that as unreadable with 0 entries while the author reads the list, and three length changes follow from it.
In 3 (gpt-oss-20b on one meeting), the matcher matched an entry naming another task or person to the one true item on shared words, so the scorer counts no true item left out and the author counts one; the author is right, and this is the kind of error check (b) measures.

\section{Data, models and licences}
\label{app:data}

Table~\ref{tab:licences} lists the nine sets with their sources and licences, and Table~\ref{tab:models} the models with their pinned revisions, requests and compute.

\begin{table*}[t]
  \centering
  \small
  \begin{adjustbox}{max width=\linewidth}
  \begin{tabular}{@{}lllll@{}}
    \toprule
    Set & One document is & Item type & Source dataset & Licence \\
    \midrule
    meetings & a window of a meeting transcript & summary sentence & AMI Meeting Corpus \citep{carletta2005ami} & CC BY 4.0 \\
    emails & one Enron email & exact quote & MailEx \citep{srivastava2023mailex} & CC BY-SA 4.0 \\
    receipts & one receipt as plain text & name & CORD v2 \citep{park2019cord} & CC BY 4.0 \\
    job-ad skills & 4--10 sentences of a job ad & exact quote & SkillSpan \citep{zhang2022skillspan} & CC BY 4.0; TECH ads CC BY-SA \\
    clinical & 3--8 sentences of a case report & name & MACCROBAT2020 \citep{caufield2019maccrobat} & CC BY 4.0 \\
    reviews & one product or service review & exact quote & ROAST \citep{roast} & Apache-2.0 (annotations) \\
    job-ad knowledge & 4--10 sentences of a job ad & name & SkillSpan & CC BY 4.0; TECH ads CC BY-SA \\
    Wikipedia people & one Wikipedia sentence & name & Few-NERD \citep{ding2021fewnerd} & CC BY-SA 4.0 \\
    novel quotes & 1--3 paragraphs of a novel & exact quote & PDNC \citep{vishnubhotla2022pdnc} & CC BY 4.0 \\
    \bottomrule
  \end{tabular}
  \end{adjustbox}
  \caption{The nine sets and their sources. SkillSpan's release is CC BY 4.0, but its paper states that the TECH job ads, from Stack Overflow, are CC BY-SA (71 of the 127 job-ad documents). Review text is not redistributed; those documents are rebuilt locally from the source.}
  \label{tab:licences}
\end{table*}

\begin{table*}[t]
  \centering
  \small
  \begin{adjustbox}{max width=\linewidth}
  \begin{tabular}{@{}lll@{}}
    \toprule
    Model & Served as & Revision or model id \\
    \midrule
    Qwen3.5-0.8B & GPU, vLLM & \texttt{Qwen/Qwen3.5-0.8B} @ \texttt{2fc06364715b} \\
    Qwen3.5-2B & GPU, vLLM & \texttt{Qwen/Qwen3.5-2B} @ \texttt{15852e8c1636} \\
    Qwen3.5-9B & GPU, vLLM & \texttt{Qwen/Qwen3.5-9B} @ \texttt{c20223623576} \\
    Qwen3.5-27B & GPU, vLLM & \texttt{Qwen/Qwen3.5-27B} @ \texttt{fc05daec18b0} \\
    Gemma 4 31B & GPU, vLLM & \texttt{google/gemma-4-31B-it} @ \texttt{842da3794eaa} \\
    Olmo 3.1 32B & GPU, vLLM & \texttt{allenai/Olmo-3.1-32B-Instruct} @ \texttt{ac0587e4a774} \\
    Ministral 3 14B & GPU, vLLM & \texttt{mistralai/Ministral-3-14B-Instruct-2512-BF16} @ \texttt{3cea74c1ebaf} \\
    gpt-oss-20b & GPU, vLLM & \texttt{openai/gpt-oss-20b} @ \texttt{6cee5e81ee83} \\
    \midrule
    DeepSeek V4.1 Flash & Together, every condition & \texttt{deepseek-ai/DeepSeek-V4.1-Flash} \\
    Qwen3.5-9B & Together, every condition & \texttt{Qwen/Qwen3.5-9B} \\
    Kimi K3 & Bedrock, prompt-side & \texttt{us.moonshotai.kimi-k3} \\
    DeepSeek V3.2 & Bedrock, prompt-side & \texttt{deepseek.v3.2} \\
    Qwen3-235B & Bedrock, prompt-side & \texttt{qwen.qwen3-235b-a22b-2507-v1:0} \\
    Mistral Large 3 & Bedrock, prompt-side & \texttt{mistral.mistral-large-3-675b-instruct} \\
    gpt-oss-120b & Bedrock, prompt-side & \texttt{openai.gpt-oss-120b-1:0} \\
    Ministral 3 14B & Bedrock, prompt-side & \texttt{mistral.ministral-3-14b-instruct} \\
    gpt-oss-20b & Bedrock, prompt-side & \texttt{openai.gpt-oss-20b-1:0} \\
    \bottomrule
  \end{tabular}
  \end{adjustbox}
  \caption{Models. GPU models are pinned to a Hugging Face revision (first 12 characters shown); hosted models are named by the service's model id and pin no revision. GPU and Together models have 26,784 request ids each and Bedrock models 23,436; the no-count prompt answers its ids at all three $k$, so 25,668 and 22,320 prompts are sent. The GPU models ran on one NVIDIA RTX PRO 6000 Blackwell (96 GB): about 31 GPU hours for the full study and 3 h 52 min for the pilot. Hosted inference cost about 6.63 USD on Together AI for the full study and 3.64 USD for the pilot, and about 58.80 USD on Amazon Bedrock.}
  \label{tab:models}
\end{table*}

\section{Answer key, $n = 0$ and generality}
\label{app:key}
\label{app:generality}

\paragraph{Blind check.}
On 40 documents from eight sets, each author counted the items alone, without seeing the answer lists.
Author A matched the list's count on 35 (87.5\%; 95\% Wilson interval 74 to 95\%), author B on 38 (95.0\%), and A and B agreed on 35 (AC1 0.851), passing the registered thresholds of 85\% and 0.80; 15 of the 40 were marked unclear by either author.
Per set (5 documents each; A matches, B matches, A = B): clinical, emails and job-ad knowledge 5, 5, 5; novel quotes, receipts and Wikipedia people 5, 5, 5; meetings 3, 5, 3; job-ad skills 2, 3, 2.
Before the first plan was frozen, an earlier round under one-line item definitions (A 72.5\%, B 65.0\%) showed which counting rules the definitions left unstated; five definitions were extended to state them, and the registered check uses the final definitions: 25 fresh documents of the five extended sets and 15 earlier documents of the three unchanged ones.
No answer list was changed after the check.
The reviews set, whose text may not be shared, was counted the same way in a separate check of five documents: author A matched the list on all five, author B on four, and the two agreed on four.
Both authors read the 60 pilot documents with their lists during the cleanup review.

\begin{table}[t]
  \centering
  \small
  \begin{adjustbox}{max width=\linewidth}
  \begin{tabular}{@{}lrrrr@{}}
    \toprule
    Set & Docs & Same & Higher & Lower \\
    \midrule
    meetings & 70 & 44 & 22 & 4 \\
    emails & 61 & 39 & 4 & 18 \\
    receipts & 55 & 55 & 0 & 0 \\
    job-ad skills & 70 & 55 & 15 & 0 \\
    clinical & 70 & 55 & 1 & 14 \\
    reviews & 55 & 21 & 9 & 25 \\
    job-ad knowledge & 57 & 46 & 5 & 6 \\
    Wikipedia people & 60 & 49 & 3 & 8 \\
    novel quotes & 60 & 59 & 0 & 1 \\
    \bottomrule
  \end{tabular}
  \end{adjustbox}
  \caption{Cleaned answer lists against the source datasets' own counts: documents whose cleaned count is the same, higher or lower.}
  \label{tab:keysource}
\end{table}

\begin{table*}[t]
  \centering
  \small
  \begin{adjustbox}{max width=\linewidth}
  \begin{tabular}{@{}lrrrrr@{}}
    \toprule
     & Frozen lists & Annotators' counts & List kept & List changed & Overlay \\
    \midrule
      C1 & +0.61 [+0.42, +0.80] & +0.65 [+0.43, +0.85] & +0.50 [+0.28, +0.70] & +0.98 [+0.55, +1.37] & +0.61 [+0.41, +0.79] \\
  C2 & +0.62 [+0.46, +0.78] & +0.63 [+0.46, +0.79] & +0.68 [+0.49, +0.88] & +0.46 [+0.20, +0.72] & +0.62 [+0.46, +0.79] \\
  C3 & +1.11 [+0.97, +1.25] & +1.13 [+0.98, +1.28] & +1.12 [+0.95, +1.30] & +1.08 [+0.87, +1.29] & +1.11 [+0.97, +1.25] \\
  C4 & +0.10 [+0.06, +0.15] & +0.09 [+0.05, +0.14] & +0.09 [+0.04, +0.15] & +0.13 [+0.07, +0.19] & +0.10 [+0.06, +0.15] \\
 
    \bottomrule
  \end{tabular}
  \end{adjustbox}
  \caption{Contrasts 1 to 4 (entries) with the answer-key variants of protocol 4.2 and 4.3. The annotators' counts change only which documents are in the $n < k$ cells. List kept and list changed split the documents by whether the cleanup changed their answer list. The overlay applies the authors' 5 corrections (Appendix~\ref{app:checks}). Contrast 2 on over-count: frozen $+20.9$ [19.5, 22.3], annotators' counts $+19.3$ [17.9, 20.7], overlay $+21.1$ [19.6, 22.6].}
  \label{tab:keyvariants}
\end{table*}

\begin{table}[t]
  \centering
  \small
  \begin{adjustbox}{max width=\linewidth}
  \begin{tabular}{@{}lrr@{}}
    \toprule
     & $n = 0$ & $n = 1$ to 4, $n < k$ \\
    \midrule
      C1 & +0.76 [+0.36, +1.16] & +0.55 [+0.34, +0.76] \\
  C2 & +0.73 [+0.42, +1.04] & +0.59 [+0.40, +0.78] \\
  C3 & +1.19 [+0.89, +1.52] & +1.09 [+0.94, +1.24] \\
  C4 & +0.01 [$-$0.11, +0.13] & +0.13 [+0.09, +0.18] \\
  C5 & $-$0.69 [$-$0.75, $-$0.63] & $-$0.49 [$-$0.54, $-$0.45] \\
 
    \bottomrule
  \end{tabular}
  \end{adjustbox}
  \caption{Confirmatory contrasts with $n = 0$ apart from $n \geq 1$ (protocol 6.1).}
  \label{tab:n0}
\end{table}

\paragraph{Generality (contrast 2).}
\begin{itemize}
  \setlength\itemsep{0pt}
    \item models whose own interval is above 0: 6 of 8 (Qwen3.5-9B, Qwen3.5-27B, Gemma 4 31B, Olmo 3.1 32B, Ministral 3 14B, gpt-oss-20b)
  \item the 5 families weighted equally (gemma, gpt-oss, mistral, olmo, qwen; within a family its models equally): +0.72 [+0.62, +0.81]
  \item leaving out gemma: +0.47 [+0.29, +0.65]
  \item leaving out gpt-oss: +0.59 [+0.40, +0.77]
  \item leaving out mistral: +0.67 [+0.48, +0.85]
  \item leaving out olmo: +0.67 [+0.49, +0.85]
  \item leaving out qwen: +0.78 [+0.69, +0.86]

\end{itemize}

\section{Pooled confirmatory tests and per set}
\label{app:perset}

Table~\ref{tab:confirmatory} gives the five confirmatory contrasts pooled with their tests, Table~\ref{tab:perset} gives them per set, and Table~\ref{tab:wordings} gives each prose wording.

\begin{table*}[t]
  \centering
  \small
  \begin{adjustbox}{max width=\linewidth}
  \begin{tabular}{@{}lrrrrr@{}}
    \toprule
    \# & Estimate [95\% CI] & $p$ & Holm $p$ & Pairs & Groups \\
    \midrule
      C1 & +0.61 [+0.42, +0.80] & $<$ 0.0001 & $<$ 0.0001 & 3000 & 366 \\
  C2 & +0.62 [+0.46, +0.78] & $<$ 0.0001 & $<$ 0.0001 & 7272 & 520 \\
  C3 & +1.11 [+0.97, +1.25] & $<$ 0.0001 & $<$ 0.0001 & 7272 & 520 \\
  C4 & +0.10 [+0.06, +0.15] & $<$ 0.0001 & $<$ 0.0001 & 7272 & 520 \\
  C5 & $-$0.54 [$-$0.58, $-$0.50] & $<$ 0.0001 & $<$ 0.0001 & 7266 & 520 \\
 
    \bottomrule
  \end{tabular}
  \end{adjustbox}

  \vspace{6pt}
  \begin{adjustbox}{max width=\linewidth}
  \begin{tabular}{@{}lrrr@{}}
    \toprule
    \# & Without pilot documents & Mixed model & Over-count (points) \\
    \midrule
      C1 & +0.54 [+0.30, +0.76] & +0.65 [+0.44, +0.87] (entries) & +0.8 [$-$0.2, +1.7] \\
  C2 & +0.66 [+0.48, +0.84] & +0.63 [+0.47, +0.79] (entries) & +20.9 [+19.5, +22.3] \\
  C3 & +1.16 [+1.01, +1.32] & +1.13 [+0.99, +1.28] (entries) & +19.7 [+18.4, +21.0] \\
  C4 & +0.11 [+0.05, +0.16] & +0.11 [+0.06, +0.16] (entries) & +8.2 [+7.3, +9.0] \\
  C5 & $-$0.55 [$-$0.59, $-$0.51] & 0.57 [0.56, 0.58] (rate ratio) & -- \\
 
    \bottomrule
  \end{tabular}
  \end{adjustbox}
  \caption{Confirmatory contrasts, eight GPU models pooled, documents with $n < k$, $k = 3$ and 5. C1 to C4 in entries, C5 in excess entries per forced slot. Mixed model (statsmodels \texttt{MixedLM}; for C5 \texttt{PoissonBayesMixedGLM}): C1 to C4 the intercept of a linear mixed model of the paired difference ($\sim n$ + model + set + words + (1 $\mid$ document)); C5 the rate ratio of the escape in a Poisson mixed model. Pairs: model--document--$k$ pairs (C1: model--document pairs); groups: the document groups the bootstrap resamples.}
  \label{tab:confirmatory}
\end{table*}

\begin{table*}[t]
  \centering
  \small
  \begin{adjustbox}{max width=\linewidth}
  \begin{tabular}{@{}lrrr@{}}
    \toprule
    Set & C1 & C2 & C3 \\
    \midrule
      meetings & +1.58 [+1.50, +1.66] & +0.19 [$-$0.30, +0.61] & +0.64 [+0.38, +0.88] \\
  emails & +0.48 [+0.31, +0.65] & +0.90 [+0.76, +1.05] & +0.93 [+0.77, +1.12] \\
  receipts & +0.88 [+0.69, +1.07] & +1.01 [+0.84, +1.20] & +1.09 [+0.91, +1.28] \\
  job-ad skills & +0.74 [$-$0.42, +1.83] & +1.54 [+0.83, +2.30] & +2.55 [+1.84, +3.31] \\
  clinical & +0.16 [$-$0.01, +0.29] & +0.28 [+0.19, +0.37] & +0.30 [+0.20, +0.40] \\
  reviews & +1.10 [+0.94, +1.24] & +0.28 [+0.08, +0.47] & +0.88 [+0.77, +0.99] \\
  job-ad knowledge & +0.45 [$-$0.74, +1.66] & +0.76 [$-$0.26, +1.79] & +2.95 [+2.06, +3.82] \\
  Wikipedia people & $-$0.17 [$-$0.26, $-$0.09] & +0.07 [+0.01, +0.15] & +0.06 [+0.00, +0.13] \\
  novel quotes & +0.55 [+0.42, +0.69] & +0.61 [+0.47, +0.77] & +0.64 [+0.49, +0.80] \\
 
    \bottomrule
  \end{tabular}
  \end{adjustbox}
  \hfill
  \begin{adjustbox}{max width=\linewidth}
  \begin{tabular}{@{}lrrr@{}}
    \toprule
    Set & C4 & C5 & C2 over-count (pts) \\
    \midrule
      meetings & +0.00 [$-$0.05, +0.05] & $-$0.46 [$-$0.58, $-$0.36] & +27.7 [+22.6, +33.3] \\
  emails & +0.16 [+0.10, +0.22] & $-$0.47 [$-$0.53, $-$0.42] & +27.7 [+22.3, +33.2] \\
  receipts & +0.38 [+0.30, +0.47] & $-$0.69 [$-$0.81, $-$0.57] & +31.2 [+26.2, +36.2] \\
  job-ad skills & +0.10 [$-$0.13, +0.34] & $-$0.61 [$-$0.72, $-$0.50] & +22.9 [+18.5, +27.4] \\
  clinical & +0.13 [+0.09, +0.17] & $-$0.57 [$-$0.64, $-$0.50] & +15.2 [+11.4, +19.1] \\
  reviews & +0.23 [+0.17, +0.29] & $-$0.07 [$-$0.11, $-$0.03] & +14.6 [+9.5, +20.2] \\
  job-ad knowledge & $-$0.34 [$-$0.60, $-$0.07] & $-$0.88 [$-$1.12, $-$0.69] & +24.4 [+21.5, +27.3] \\
  Wikipedia people & +0.05 [+0.02, +0.08] & $-$0.62 [$-$0.71, $-$0.55] & +6.5 [+4.6, +8.7] \\
  novel quotes & +0.30 [+0.25, +0.36] & $-$0.40 [$-$0.48, $-$0.32] & +18.7 [+15.6, +21.9] \\
 
    \bottomrule
  \end{tabular}
  \end{adjustbox}
  \caption{Confirmatory contrasts per set, eight GPU models pooled: C1 to C4 in entries, C5 in excess entries per forced slot, the last column in over-count points. Job-ad skills is the least certain answer list.}
  \label{tab:perset}
\end{table*}

\begin{table}[t]
  \centering
  \small
  \setlength{\tabcolsep}{3pt}
  \begin{adjustbox}{max width=\linewidth}
  \begin{tabular}{@{}lrr@{}}
    \toprule
    Wording (emails) & vs \cnd{none} & Over-count \% \\
    \midrule
      \multicolumn{3}{@{}l}{\cnd{prose\_up\_to\_k}} \\
  Extract up to $k$ requests from this document. & $-$0.85 [$-$1.02, $-$0.69] & 45.2 \\
  List up to $k$ requests. & $-$0.79 [$-$0.97, $-$0.61] & 42.4 \\
  You may include up to $k$ requests. & $-$0.61 [$-$0.77, $-$0.45] & 40.2 \\
  Return maximum $k$ requests. & $-$0.84 [$-$1.02, $-$0.67] & 45.9 \\
  Do not include more than $k$ requests. & $-$0.72 [$-$0.89, $-$0.56] & 40.0 \\
  \midrule
  \multicolumn{3}{@{}l}{\cnd{prose\_exactly\_k}} \\
  Extract exactly $k$ requests. & $-$0.64 [$-$0.81, $-$0.47] & 59.7 \\
  Extract $k$ requests. & $-$0.51 [$-$0.68, $-$0.33] & 53.7 \\
  List $k$ requests. & $-$0.43 [$-$0.61, $-$0.25] & 51.0 \\
  Return the top $k$ requests. & $-$0.71 [$-$0.89, $-$0.52] & 44.2 \\
  Extract the first $k$ requests. & $-$1.02 [$-$1.19, $-$0.84] & 45.8 \\
 
    \bottomrule
  \end{tabular}
  \end{adjustbox}
  \caption{Prose wordings, shown for emails (other sets name their own item), with the length change against \cnd{none} (entries) and over-count, pooled over all sets and the eight GPU models, $n < k$, $k = 3$ and 5. The schema conditions use ``List all the requests.'' (the escape and ``fewer is fine'' add their line). Templates and the real prompts they are modelled on: \texttt{data/wordings.yaml}.}
  \label{tab:wordings}
\end{table}

\section{Named tests per model}
\label{app:named}

\begin{table*}[t]
  \centering
  \small
  \begin{adjustbox}{max width=\linewidth}
  \begin{tabular}{@{}lrrr@{}}
    \toprule
    Model & \cnd{fewer\_ok} vs \cnd{none} & \cnd{fewer\_ok} vs \cnd{min\_k\_shown} & \cnd{max\_k\_shown} vs \cnd{none} \\
    \midrule
      Qwen3.5-0.8B & +0.03 [$-$0.39, +0.47] & +0.62 [+0.19, +1.05] & $-$0.27 [$-$0.64, +0.09] \\
  Qwen3.5-2B & $-$2.70 [$-$3.91, $-$1.56] & $-$3.31 [$-$4.26, $-$2.35] & $-$1.52 [$-$2.37, $-$0.70] \\
  Qwen3.5-9B & $-$0.58 [$-$0.70, $-$0.46] & $-$1.20 [$-$1.34, $-$1.06] & $-$0.19 [$-$0.27, $-$0.10] \\
  Qwen3.5-27B & +0.46 [+0.37, +0.56] & $-$0.78 [$-$0.88, $-$0.68] & $-$0.05 [$-$0.12, +0.02] \\
  Gemma 4 31B & +0.02 [$-$0.02, +0.06] & $-$1.65 [$-$1.79, $-$1.52] & $-$0.02 [$-$0.05, +0.01] \\
  Olmo 3.1 32B & $-$1.92 [$-$2.20, $-$1.66] & $-$2.18 [$-$2.43, $-$1.95] & $-$0.94 [$-$1.19, $-$0.70] \\
  Ministral 3 14B & $-$1.50 [$-$1.76, $-$1.25] & $-$1.80 [$-$1.99, $-$1.62] & $-$0.83 [$-$1.02, $-$0.65] \\
  gpt-oss-20b & +0.06 [$-$0.08, +0.19] & $-$0.81 [$-$0.93, $-$0.69] & $-$0.11 [$-$0.26, +0.03] \\
 
    \bottomrule
  \end{tabular}
  \end{adjustbox}

  \vspace{6pt}
  \begin{adjustbox}{max width=\linewidth}
  \begin{tabular}{@{}lrrr@{}}
    \toprule
    Model & $n \geq k$: shown vs \cnd{none} & $n < k$ minus $n \geq k$ ($k = 3, 5$) & Empty, $k = 1$: shown vs \cnd{none} (pts) \\
    \midrule
      Qwen3.5-0.8B & $-$0.25 [$-$0.60, +0.11] & +0.44 [$-$0.38, +1.29] & $-$1.0 [$-$3.8, +1.9] \\
  Qwen3.5-2B & $-$0.43 [$-$1.46, +0.52] & +0.55 [$-$1.42, +2.58] & +1.9 [+0.0, +4.8] \\
  Qwen3.5-9B & $-$0.00 [$-$0.13, +0.12] & +0.72 [+0.33, +1.03] & +6.7 [+2.9, +11.4] \\
  Qwen3.5-27B & $-$0.09 [$-$0.16, $-$0.03] & +1.52 [+1.31, +1.75] & +7.6 [+2.9, +12.4] \\
  Gemma 4 31B & +0.08 [+0.05, +0.12] & +1.48 [+1.34, +1.63] & +15.2 [+9.5, +21.0] \\
  Olmo 3.1 32B & $-$0.24 [$-$0.37, $-$0.12] & +0.89 [+0.63, +1.17] & +4.8 [$-$1.0, +10.5] \\
  Ministral 3 14B & $-$0.32 [$-$0.45, $-$0.19] & +1.01 [+0.72, +1.32] & +3.8 [$-$1.0, +8.6] \\
  gpt-oss-20b & +0.05 [$-$0.06, +0.17] & +0.80 [+0.54, +1.04] & +20.0 [+14.3, +25.7] \\
 
    \bottomrule
  \end{tabular}
  \end{adjustbox}
  \caption{Named tests per GPU model (entries unless marked).}
  \label{tab:namedpermodel}
\end{table*}

By $k$, the shown minimum against no count on documents with $n \geq k$ is $-0.08$ [$-0.20$, $+0.04$] at $k = 1$, $-0.25$ [$-0.51$, $+0.01$] at $k = 3$ and $-0.71$ [$-1.57$, $-0.02$] at $k = 5$; on $n < k$ it is $+0.22$, $+0.28$ and $+0.86$, and the differences ($n < k$ minus $n \geq k$) are $+0.30$ [$-0.06$, $+0.68$], $+0.54$ [0.20, 0.87] and $+1.57$ [0.84, 2.58].

\section{Outcomes per condition and model}
\label{app:outcomes}

Tables~\ref{tab:outcomes} and~\ref{tab:outcomespooled} give the outcomes by condition, pooled; the tables after them give the same per model, and Table~\ref{tab:unreadable} counts the unreadable outputs.

\begin{table*}[t]
  \centering
  \small
  \begin{adjustbox}{max width=\linewidth}
  \begin{tabular}{@{}lrrrrrrrr@{}}
    \toprule
     & \multicolumn{4}{c}{Sorting, $n < k$ (\%)} & \multicolumn{3}{c}{Length against $k$, $n < k$ (\%)} & Left out\pre{} \\
    \cmidrule(lr){2-5}\cmidrule(lr){6-8}
    Condition & padded & says short & short, no note & unreadable & fewer than $k$ & exactly $k$ & more than $k$ & (\%, $n \geq 1$) \\
    \midrule
      \cnd{none} & 34.5 & 0.3 & 58.3 & 6.9 & 70.0 & 5.6 & 17.4 & 17.0 \\
  \cnd{prose\_up\_to\_k} & 42.7 & 0.1 & 53.5 & 3.7 & 65.6 & 24.0 & 6.6 & 20.0 \\
  \cnd{prose\_exactly\_k} & 50.9 & 0.6 & 42.5 & 6.0 & 50.1 & 39.0 & 4.9 & 21.6 \\
  \cnd{max\_k\_shown} & 35.7 & 0.2 & 56.7 & 7.3 & 68.0 & 15.5 & 9.2 & 20.9 \\
  \cnd{min\_k\_shown} & 55.4 & 5.6 & 28.5 & 10.5 & 37.1 & 35.5 & 17.0 & 18.2 \\
  \cnd{min\_k\_shown\_fewer\_ok} & 31.0 & 2.0 & 59.7 & 7.3 & 67.4 & 15.0 & 10.2 & 24.3 \\
  \cnd{min\_k\_enforced} & 90.1 & 4.0 & 5.7 & 0.2 & 0.6 & 71.8 & 27.5 & 8.6 \\
  \cnd{min\_k\_escape} & 51.0 & 48.0 & 0.9 & 0.1 & 0.2 & 78.5 & 21.2 & 13.6 \\
 
    \bottomrule
  \end{tabular}
  \end{adjustbox}
  \caption{What the outputs are, by condition, eight GPU models pooled, $k = 3$ and 5. Padded outputs have more substantive entries than $n$; this wins over a note. Left out: share of the true items of documents with $n \geq 1$ that no entry matches. Per model: Tables~\ref{tab:outcomesqwen} and~\ref{tab:outcomesother}.}
  \label{tab:outcomes}
\end{table*}

\begin{table*}[t]
  \centering
  \small
  \begin{adjustbox}{max width=\linewidth}\begin{tabular}{lrrrrrr}
  \toprule
  Condition & Over-count \% & Bound \% & Excess & Unreadable \% & Exactly $k$ \% & Left out \% \\
  \midrule
  \cnd{none} & 34.5 & 34.3 & 2.41 & 6.9 & 5.6 & 17.0 \\
  \cnd{prose\_up\_to\_k} & 42.7 & 42.4 & 1.58 & 3.7 & 24.0 & 20.0 \\
  \cnd{prose\_exactly\_k} & 50.9 & 50.6 & 1.71 & 6.0 & 39.0 & 21.6 \\
  \cnd{max\_k\_shown} & 35.7 & 35.5 & 1.92 & 7.3 & 15.5 & 20.9 \\
  \cnd{min\_k\_shown} & 55.4 & 55.2 & 3.02 & 10.5 & 35.5 & 18.2 \\
  \cnd{min\_k\_shown\_fewer\_ok} & 31.0 & 30.8 & 1.76 & 7.3 & 15.0 & 24.3 \\
  \cnd{min\_k\_enforced} & 90.1 & 89.5 & 3.49 & 0.2 & 71.8 & 8.6 \\
  \cnd{min\_k\_escape} & 51.0 & 50.6 & 2.00 & 0.1 & 78.5 & 13.6 \\
  \bottomrule
\end{tabular}
\end{adjustbox}
  \caption{Over-count, its placeholder bound (every substantive entry of 8 words or fewer holding ``no'', ``none'', ``not'', ``only'', ``other'', ``additional'' or ``found'' taken as a note), mean excess, unreadable share, exactly $k$, on $n < k$, and true items left out\pre{} on $n \geq 1$, eight GPU models pooled, $k = 3$ and 5.}
  \label{tab:outcomespooled}
\end{table*}

\begin{table*}[p]
  \centering
  \small
  \setlength{\tabcolsep}{4pt}
  \begin{adjustbox}{max width=\linewidth}
  \begin{tabular}{@{}lrrrrrrrrrr@{}}
    \toprule
    Condition & Outputs & Over \% & Bound \% & Excess & Unread. \% & $< k$ \% & $= k$ \% & $> k$ \% & $= n$ \% & Left out\pre{} \% \\
    \midrule
      \multicolumn{11}{@{}l}{\textit{Qwen3.5-0.8B}} \\
  \cnd{none} & 909 & 30.9 & 30.9 & 3.45 & 54.2 & 19.7 & 1.8 & 24.3 & 9.5 & 66.7 \\
  \cnd{prose\_up\_to\_k} & 4545 & 50.0 & 49.8 & 2.75 & 28.0 & 32.5 & 20.2 & 19.4 & 16.3 & 53.5 \\
  \cnd{prose\_exactly\_k} & 4545 & 39.6 & 39.4 & 2.35 & 45.0 & 21.0 & 18.6 & 15.3 & 11.8 & 65.1 \\
  \cnd{max\_k\_shown} & 909 & 29.3 & 29.2 & 3.21 & 57.5 & 18.0 & 2.2 & 22.2 & 9.6 & 68.8 \\
  \cnd{min\_k\_shown} & 909 & 24.9 & 24.9 & 3.02 & 67.4 & 11.6 & 2.4 & 18.6 & 5.6 & 76.5 \\
  \cnd{min\_k\_shown\_fewer\_ok} & 909 & 34.0 & 33.9 & 3.42 & 49.1 & 21.2 & 3.7 & 26.0 & 12.9 & 65.1 \\
  \cnd{min\_k\_enforced} & 909 & 91.9 & 91.5 & 6.13 & 0.8 & 1.0 & 34.3 & 63.9 & 6.6 & 15.6 \\
  \cnd{min\_k\_escape} & 908 & 89.6 & 89.1 & 5.32 & 1.0 & 0.8 & 35.0 & 63.2 & 7.9 & 15.1 \\
  \midrule
  \multicolumn{11}{@{}l}{\textit{Qwen3.5-2B}} \\
  \cnd{none} & 909 & 55.1 & 55.0 & 8.97 & 0.1 & 56.8 & 5.9 & 37.2 & 38.3 & 15.1 \\
  \cnd{prose\_up\_to\_k} & 4545 & 58.9 & 58.7 & 4.36 & 0.6 & 51.4 & 26.4 & 21.5 & 34.7 & 21.1 \\
  \cnd{prose\_exactly\_k} & 4545 & 58.4 & 58.3 & 3.92 & 0.2 & 50.8 & 30.4 & 18.7 & 35.8 & 21.8 \\
  \cnd{max\_k\_shown} & 909 & 55.7 & 55.3 & 7.45 & 0.1 & 55.7 & 9.4 & 34.9 & 37.3 & 17.5 \\
  \cnd{min\_k\_shown} & 909 & 60.4 & 60.3 & 9.56 & 0.2 & 47.9 & 14.6 & 37.3 & 34.1 & 17.0 \\
  \cnd{min\_k\_shown\_fewer\_ok} & 909 & 46.9 & 46.9 & 6.47 & 0.7 & 56.0 & 9.6 & 33.8 & 34.3 & 23.8 \\
  \cnd{min\_k\_enforced} & 908 & 94.5 & 94.1 & 4.65 & 0.0 & 1.1 & 43.7 & 55.2 & 5.2 & 12.1 \\
  \cnd{min\_k\_escape} & 907 & 83.0 & 82.8 & 3.99 & 0.1 & 0.8 & 48.8 & 50.3 & 13.9 & 13.6 \\
  \midrule
  \multicolumn{11}{@{}l}{\textit{Qwen3.5-9B}} \\
  \cnd{none} & 909 & 26.3 & 26.1 & 0.98 & 0.0 & 84.4 & 5.2 & 10.5 & 61.5 & 13.5 \\
  \cnd{prose\_up\_to\_k} & 4545 & 34.1 & 33.7 & 0.81 & 0.1 & 79.3 & 18.7 & 1.9 & 56.9 & 17.8 \\
  \cnd{prose\_exactly\_k} & 4545 & 38.9 & 38.6 & 0.92 & 0.0 & 71.2 & 27.5 & 1.4 & 51.5 & 18.2 \\
  \cnd{max\_k\_shown} & 909 & 28.9 & 28.5 & 0.79 & 0.0 & 84.0 & 11.6 & 4.4 & 58.7 & 16.6 \\
  \cnd{min\_k\_shown} & 909 & 44.8 & 44.6 & 1.56 & 0.1 & 62.0 & 25.7 & 12.1 & 46.1 & 11.6 \\
  \cnd{min\_k\_shown\_fewer\_ok} & 909 & 18.0 & 17.9 & 0.67 & 0.0 & 86.4 & 6.6 & 7.0 & 55.1 & 26.9 \\
  \cnd{min\_k\_enforced} & 909 & 92.7 & 92.4 & 3.24 & 0.2 & 0.7 & 75.4 & 23.8 & 6.6 & 6.4 \\
  \cnd{min\_k\_escape} & 909 & 47.2 & 46.9 & 1.40 & 0.0 & 0.0 & 87.5 & 12.5 & 40.2 & 13.2 \\
  \midrule
  \multicolumn{11}{@{}l}{\textit{Qwen3.5-27B}} \\
  \cnd{none} & 909 & 23.1 & 23.1 & 0.52 & 0.0 & 88.6 & 5.8 & 5.6 & 71.0 & 5.2 \\
  \cnd{prose\_up\_to\_k} & 4545 & 29.5 & 29.5 & 0.59 & 0.7 & 80.7 & 18.4 & 0.1 & 64.9 & 10.0 \\
  \cnd{prose\_exactly\_k} & 4545 & 42.1 & 42.0 & 0.94 & 0.8 & 64.2 & 34.9 & 0.1 & 53.6 & 9.8 \\
  \cnd{max\_k\_shown} & 909 & 26.4 & 26.3 & 0.46 & 0.0 & 85.9 & 14.1 & 0.0 & 68.6 & 9.5 \\
  \cnd{min\_k\_shown} & 909 & 60.9 & 60.6 & 1.72 & 2.4 & 38.9 & 51.4 & 7.3 & 35.0 & 4.8 \\
  \cnd{min\_k\_shown\_fewer\_ok} & 909 & 42.2 & 42.1 & 0.97 & 1.5 & 61.9 & 35.4 & 1.1 & 51.6 & 7.1 \\
  \cnd{min\_k\_enforced} & 909 & 91.6 & 91.4 & 2.57 & 0.0 & 0.6 & 91.2 & 8.3 & 7.9 & 4.1 \\
  \cnd{min\_k\_escape} & 909 & 43.6 & 43.1 & 0.97 & 0.0 & 0.1 & 97.2 & 2.6 & 53.8 & 6.2 \\
 
    \bottomrule
  \end{tabular}
  \end{adjustbox}
  \caption{Per GPU model (Qwen3.5 sizes): outputs on $n < k$, over-count, placeholder bound, excess, unreadable, fewer than $k$, exactly $k$, more than $k$, exactly $n$ (all on $n < k$), and true items left out on $n \geq 1$.}
  \label{tab:outcomesqwen}
\end{table*}

\begin{table*}[p]
  \centering
  \small
  \setlength{\tabcolsep}{4pt}
  \begin{adjustbox}{max width=\linewidth}
  \begin{tabular}{@{}lrrrrrrrrrr@{}}
    \toprule
    Condition & Outputs & Over \% & Bound \% & Excess & Unread. \% & $< k$ \% & $= k$ \% & $> k$ \% & $= n$ \% & Left out\pre{} \% \\
    \midrule
      \multicolumn{11}{@{}l}{\textit{Gemma 4 31B}} \\
  \cnd{none} & 909 & 14.4 & 14.2 & 0.20 & 0.0 & 95.6 & 3.2 & 1.2 & 77.8 & 8.6 \\
  \cnd{prose\_up\_to\_k} & 4545 & 16.5 & 16.2 & 0.24 & 0.0 & 95.1 & 4.9 & 0.0 & 76.5 & 12.0 \\
  \cnd{prose\_exactly\_k} & 4545 & 50.9 & 50.5 & 1.12 & 0.9 & 56.3 & 42.8 & 0.0 & 46.4 & 9.8 \\
  \cnd{max\_k\_shown} & 909 & 13.6 & 13.4 & 0.18 & 0.0 & 95.4 & 4.5 & 0.1 & 79.0 & 12.0 \\
  \cnd{min\_k\_shown} & 909 & 62.5 & 62.4 & 1.82 & 4.0 & 38.2 & 44.8 & 13.1 & 33.4 & 5.2 \\
  \cnd{min\_k\_shown\_fewer\_ok} & 909 & 16.0 & 15.7 & 0.22 & 0.0 & 93.8 & 5.4 & 0.8 & 77.2 & 7.9 \\
  \cnd{min\_k\_enforced} & 909 & 85.6 & 85.3 & 2.38 & 0.0 & 0.1 & 86.7 & 13.2 & 13.5 & 4.2 \\
  \cnd{min\_k\_escape} & 909 & 29.9 & 29.9 & 0.62 & 0.0 & 0.0 & 96.6 & 3.4 & 62.9 & 7.3 \\
  \midrule
  \multicolumn{11}{@{}l}{\textit{Olmo 3.1 32B}} \\
  \cnd{none} & 909 & 42.2 & 41.7 & 2.01 & 0.9 & 67.8 & 7.5 & 23.9 & 50.4 & 11.5 \\
  \cnd{prose\_up\_to\_k} & 4545 & 49.2 & 48.8 & 1.32 & 0.3 & 61.3 & 33.0 & 5.4 & 45.3 & 18.5 \\
  \cnd{prose\_exactly\_k} & 4545 & 54.8 & 54.4 & 1.34 & 0.4 & 51.6 & 46.3 & 1.7 & 39.2 & 20.2 \\
  \cnd{max\_k\_shown} & 909 & 45.0 & 44.7 & 1.07 & 0.9 & 64.4 & 31.7 & 3.1 & 47.4 & 18.4 \\
  \cnd{min\_k\_shown} & 909 & 57.8 & 57.4 & 2.25 & 1.3 & 44.2 & 37.4 & 17.1 & 35.9 & 12.9 \\
  \cnd{min\_k\_shown\_fewer\_ok} & 909 & 18.7 & 18.5 & 0.49 & 2.9 & 83.3 & 10.0 & 3.9 & 51.3 & 34.6 \\
  \cnd{min\_k\_enforced} & 909 & 95.6 & 93.0 & 3.65 & 0.0 & 0.6 & 73.2 & 26.3 & 4.2 & 9.1 \\
  \cnd{min\_k\_escape} & 909 & 49.5 & 49.3 & 1.90 & 0.0 & 0.0 & 80.5 & 19.5 & 41.9 & 15.3 \\
  \midrule
  \multicolumn{11}{@{}l}{\textit{Ministral 3 14B}} \\
  \cnd{none} & 909 & 50.2 & 49.9 & 2.07 & 0.0 & 68.0 & 7.5 & 24.5 & 47.5 & 6.9 \\
  \cnd{prose\_up\_to\_k} & 4545 & 59.9 & 59.6 & 1.51 & 0.1 & 58.0 & 39.0 & 2.9 & 38.5 & 12.5 \\
  \cnd{prose\_exactly\_k} & 4545 & 67.5 & 67.1 & 1.70 & 0.2 & 45.4 & 52.9 & 1.5 & 31.2 & 12.7 \\
  \cnd{max\_k\_shown} & 909 & 46.5 & 46.5 & 1.25 & 0.0 & 70.0 & 21.5 & 8.6 & 50.6 & 10.3 \\
  \cnd{min\_k\_shown} & 909 & 67.7 & 67.2 & 2.35 & 0.0 & 42.1 & 40.4 & 17.5 & 31.8 & 8.1 \\
  \cnd{min\_k\_shown\_fewer\_ok} & 909 & 32.7 & 32.6 & 0.72 & 0.0 & 77.9 & 19.8 & 2.3 & 55.8 & 17.2 \\
  \cnd{min\_k\_enforced} & 909 & 97.2 & 96.9 & 3.14 & 0.3 & 0.3 & 82.2 & 17.2 & 2.2 & 9.9 \\
  \cnd{min\_k\_escape} & 909 & 24.3 & 24.2 & 0.60 & 0.0 & 0.0 & 93.6 & 6.4 & 49.6 & 29.4 \\
  \midrule
  \multicolumn{11}{@{}l}{\textit{gpt-oss-20b}} \\
  \cnd{none} & 909 & 33.9 & 33.8 & 1.04 & 0.0 & 79.6 & 8.1 & 12.2 & 59.7 & 8.6 \\
  \cnd{prose\_up\_to\_k} & 4545 & 43.7 & 43.2 & 1.08 & 0.0 & 66.3 & 31.7 & 1.9 & 52.1 & 14.5 \\
  \cnd{prose\_exactly\_k} & 4545 & 54.9 & 54.5 & 1.36 & 0.3 & 40.6 & 58.5 & 0.7 & 41.2 & 15.0 \\
  \cnd{max\_k\_shown} & 909 & 40.3 & 40.0 & 0.91 & 0.0 & 70.6 & 29.4 & 0.0 & 55.2 & 14.0 \\
  \cnd{min\_k\_shown} & 909 & 64.2 & 64.0 & 1.90 & 8.1 & 11.8 & 67.2 & 12.9 & 26.2 & 9.0 \\
  \cnd{min\_k\_shown\_fewer\_ok} & 909 & 39.3 & 38.9 & 1.14 & 4.4 & 58.7 & 29.8 & 7.0 & 49.7 & 11.4 \\
  \cnd{min\_k\_enforced} & 908 & 71.8 & 71.5 & 2.13 & 0.2 & 0.1 & 87.6 & 12.1 & 27.2 & 7.3 \\
  \cnd{min\_k\_escape} & 907 & 40.6 & 39.9 & 1.21 & 0.0 & 0.0 & 88.4 & 11.6 & 53.8 & 8.7 \\
 
    \bottomrule
  \end{tabular}
  \end{adjustbox}
  \caption{As Table~\ref{tab:outcomesqwen}, for Gemma 4 31B, Olmo 3.1 32B, Ministral 3 14B and gpt-oss-20b.}
  \label{tab:outcomesother}
\end{table*}

\begin{table}[t]
  \centering
  \small
  \setlength{\tabcolsep}{3pt}
  \begin{adjustbox}{max width=\linewidth}
  \begin{tabular}{@{}lrrrr@{}}
    \toprule
    Condition & Padded & Says short & Short, no note & Unread. \\
    \midrule
      \multicolumn{5}{@{}l}{\textit{Qwen3.5-0.8B}} \\
  \cnd{none} & 30.9 & 1.1 & 13.8 & 54.2 \\
  \cnd{min\_k\_shown} & 24.9 & 0.3 & 7.4 & 67.4 \\
  \cnd{min\_k\_shown\_fewer\_ok} & 34.0 & 0.4 & 16.5 & 49.1 \\
  \cnd{max\_k\_shown} & 29.3 & 0.2 & 13.0 & 57.5 \\
  \cnd{min\_k\_enforced} & 91.9 & 3.3 & 4.1 & 0.8 \\
  \midrule
  \multicolumn{5}{@{}l}{\textit{Qwen3.5-2B}} \\
  \cnd{none} & 55.1 & 0.0 & 44.8 & 0.1 \\
  \cnd{min\_k\_shown} & 60.4 & 0.0 & 39.4 & 0.2 \\
  \cnd{min\_k\_shown\_fewer\_ok} & 46.9 & 0.0 & 52.5 & 0.7 \\
  \cnd{max\_k\_shown} & 55.7 & 0.0 & 44.2 & 0.1 \\
  \cnd{min\_k\_enforced} & 94.5 & 1.1 & 4.4 & 0.0 \\
  \midrule
  \multicolumn{5}{@{}l}{\textit{Qwen3.5-9B}} \\
  \cnd{none} & 26.3 & 0.0 & 73.7 & 0.0 \\
  \cnd{min\_k\_shown} & 44.8 & 0.0 & 55.1 & 0.1 \\
  \cnd{min\_k\_shown\_fewer\_ok} & 18.0 & 0.0 & 82.0 & 0.0 \\
  \cnd{max\_k\_shown} & 28.9 & 0.0 & 71.1 & 0.0 \\
  \cnd{min\_k\_enforced} & 92.7 & 2.2 & 4.8 & 0.2 \\
  \midrule
  \multicolumn{5}{@{}l}{\textit{Qwen3.5-27B}} \\
  \cnd{none} & 23.1 & 0.0 & 76.9 & 0.0 \\
  \cnd{min\_k\_shown} & 60.9 & 0.0 & 36.6 & 2.4 \\
  \cnd{min\_k\_shown\_fewer\_ok} & 42.2 & 0.1 & 56.1 & 1.5 \\
  \cnd{max\_k\_shown} & 26.4 & 0.0 & 73.6 & 0.0 \\
  \cnd{min\_k\_enforced} & 91.6 & 2.2 & 6.2 & 0.0 \\
 
    \bottomrule
  \end{tabular}
  \end{adjustbox}
  \caption{Sorting per GPU model (\%), Qwen3.5 sizes, selected conditions, $n < k$, $k = 3$ and 5.}
  \label{tab:sortqwen}
\end{table}

\begin{table}[t]
  \centering
  \small
  \setlength{\tabcolsep}{3pt}
  \begin{adjustbox}{max width=\linewidth}
  \begin{tabular}{@{}lrrrr@{}}
    \toprule
    Condition & Padded & Says short & Short, no note & Unread. \\
    \midrule
      \multicolumn{5}{@{}l}{\textit{Gemma 4 31B}} \\
  \cnd{none} & 14.4 & 0.0 & 85.6 & 0.0 \\
  \cnd{min\_k\_shown} & 62.5 & 18.5 & 15.1 & 4.0 \\
  \cnd{min\_k\_shown\_fewer\_ok} & 16.0 & 0.0 & 84.0 & 0.0 \\
  \cnd{max\_k\_shown} & 13.6 & 0.0 & 86.4 & 0.0 \\
  \cnd{min\_k\_enforced} & 85.6 & 8.0 & 6.4 & 0.0 \\
  \midrule
  \multicolumn{5}{@{}l}{\textit{Olmo 3.1 32B}} \\
  \cnd{none} & 42.2 & 1.5 & 55.3 & 0.9 \\
  \cnd{min\_k\_shown} & 57.8 & 9.5 & 31.5 & 1.3 \\
  \cnd{min\_k\_shown\_fewer\_ok} & 18.7 & 6.9 & 71.5 & 2.9 \\
  \cnd{max\_k\_shown} & 45.0 & 1.7 & 52.5 & 0.9 \\
  \cnd{min\_k\_enforced} & 95.6 & 4.4 & 0.0 & 0.0 \\
  \midrule
  \multicolumn{5}{@{}l}{\textit{Ministral 3 14B}} \\
  \cnd{none} & 50.2 & 0.0 & 49.8 & 0.0 \\
  \cnd{min\_k\_shown} & 67.7 & 14.1 & 18.3 & 0.0 \\
  \cnd{min\_k\_shown\_fewer\_ok} & 32.7 & 7.6 & 59.7 & 0.0 \\
  \cnd{max\_k\_shown} & 46.5 & 0.0 & 53.5 & 0.0 \\
  \cnd{min\_k\_enforced} & 97.2 & 2.0 & 0.4 & 0.3 \\
  \midrule
  \multicolumn{5}{@{}l}{\textit{gpt-oss-20b}} \\
  \cnd{none} & 33.9 & 0.0 & 66.1 & 0.0 \\
  \cnd{min\_k\_shown} & 64.2 & 2.5 & 25.1 & 8.1 \\
  \cnd{min\_k\_shown\_fewer\_ok} & 39.3 & 0.7 & 55.7 & 4.4 \\
  \cnd{max\_k\_shown} & 40.3 & 0.0 & 59.7 & 0.0 \\
  \cnd{min\_k\_enforced} & 71.8 & 9.0 & 18.9 & 0.2 \\
 
    \bottomrule
  \end{tabular}
  \end{adjustbox}
  \caption{As Table~\ref{tab:sortqwen}, for the other four GPU models.}
  \label{tab:sortother}
\end{table}

\begin{table*}[t]
  \centering
  \small
  \begin{adjustbox}{max width=\linewidth}
  \begin{tabular}{@{}lrrrrr@{}}
    \toprule
    Model & OK outputs & Unreadable & Unreadable \% & JSON, no answer list & Cut at token limit \\
    \midrule
      Qwen3.5-0.8B & 26783 & 9810 & 36.6 & 9521 & 21 \\
  Qwen3.5-2B & 26781 & 93 & 0.3 & 0 & 3 \\
  Qwen3.5-9B & 26784 & 10 & 0.0 & 2 & 2 \\
  Qwen3.5-27B & 26784 & 122 & 0.5 & 0 & 5 \\
  Gemma 4 31B & 26784 & 77 & 0.3 & 0 & 0 \\
  Olmo 3.1 32B & 26784 & 178 & 0.7 & 85 & 18 \\
  Ministral 3 14B & 26784 & 21 & 0.1 & 0 & 3 \\
  gpt-oss-20b & 26778 & 128 & 0.5 & 0 & 4 \\
  DeepSeek V4.1 Flash (Together) & 26784 & 0 & 0.0 & 0 & 0 \\
  Qwen3.5-9B (Together) & 26784 & 8 & 0.0 & 1 & 1 \\
  Kimi K3 (Bedrock) & 23436 & 0 & 0.0 & 0 & 0 \\
  DeepSeek V3.2 (Bedrock) & 23436 & 13 & 0.1 & 3 & 0 \\
  Qwen3-235B (Bedrock) & 23436 & 0 & 0.0 & 0 & 0 \\
  Mistral Large 3 (Bedrock) & 23436 & 80 & 0.3 & 0 & 80 \\
  Ministral 3 14B (Bedrock) & 23436 & 20 & 0.1 & 1 & 0 \\
  gpt-oss-20b (Bedrock) & 23436 & 127 & 0.5 & 0 & 5 \\
  gpt-oss-120b (Bedrock) & 23436 & 4 & 0.0 & 0 & 0 \\
 
    \bottomrule
  \end{tabular}
  \end{adjustbox}
  \caption{Unreadable outputs, all conditions and $k$. ``JSON, no answer list'': the parser found JSON but no list it may take as the answer, almost always the answer written inside a copy of the schema.}
  \label{tab:unreadable}
\end{table*}

\section{What the extra entries are}
\label{app:fill}

Table~\ref{tab:fillfull} gives every label the matcher assigns to the extra entries, with the placeholder and repeat shares.

\begin{table*}[t]
  \centering
  \small
  \begin{adjustbox}{max width=\linewidth}
  \begin{tabular}{@{}lrrrrrrr@{}}
    \toprule
    Condition & Extras & Split \% & Wrong kind \% & Invented \% & Echo or note \% & Placeholder \% & Repeat \% \\
    \midrule
      \cnd{none} & 18109 & 5.5 & 64.6 & 29.8 & 0.0 & 0.0 & 12.5 \\
  \cnd{max\_k\_shown} & 14697 & 5.7 & 66.3 & 27.9 & 0.1 & 0.1 & 17.3 \\
  \cnd{min\_k\_shown} & 22665 & 7.5 & 62.9 & 29.5 & 0.1 & 0.7 & 13.9 \\
  \cnd{min\_k\_shown\_fewer\_ok} & 13343 & 6.3 & 72.0 & 21.6 & 0.1 & 0.1 & 16.2 \\
  \cnd{min\_k\_enforced} & 26193 & 9.6 & 79.6 & 9.9 & 0.9 & 5.8 & 16.1 \\
  \cnd{min\_k\_escape} & 15198 & 9.8 & 82.4 & 7.2 & 0.6 & 34.2 & 13.3 \\
 
    \bottomrule
  \end{tabular}
  \end{adjustbox}
  \caption{Matcher labels\pre{} of the extra entries, $n < k$, $k = 3$ and 5, eight GPU models pooled. ``Echo or note'': entries the matcher would call invented that copy the prompt or may be a note. Placeholder and repeat shares are of all list entries and come from the frozen placeholder rules, not the matcher.}
  \label{tab:fillfull}
\end{table*}

\section{Hosted reference models}
\label{app:hosted}

The hosted models are a reference and never pooled with the GPU models: Table~\ref{tab:hosted} gives their contrasts, Table~\ref{tab:hostedsorting} their sorting and Table~\ref{tab:pairs} the three same-weights pairs.

\begin{table*}[t]
  \centering
  \small
  \begin{adjustbox}{max width=\linewidth}
  \begin{tabular}{@{}lrrr@{}}
    \toprule
    Model & C1 & C2 & C3 \\
    \midrule
      DeepSeek V4.1 Flash (Together) & +1.23 [+1.05, +1.43] & +1.40 [+1.25, +1.56] & +1.60 [+1.45, +1.75] \\
  Qwen3.5-9B (Together) & +0.50 [+0.32, +0.68] & +0.64 [+0.50, +0.78] & +0.84 [+0.73, +0.96] \\
  Kimi K3 (Bedrock) & +0.77 [+0.64, +0.91] & +1.25 [+1.12, +1.38] & +1.21 [+1.09, +1.33] \\
  DeepSeek V3.2 (Bedrock) & +1.30 [+1.07, +1.54] & +2.39 [+2.19, +2.59] & +2.19 [+2.01, +2.38] \\
  Qwen3-235B (Bedrock) & +1.50 [+1.34, +1.66] & +1.28 [+1.13, +1.43] & +1.35 [+1.23, +1.48] \\
  Mistral Large 3 (Bedrock) & +1.01 [+0.87, +1.15] & +0.98 [+0.82, +1.14] & +1.24 [+1.12, +1.38] \\
  Ministral 3 14B (Bedrock) & +0.99 [+0.80, +1.20] & +0.28 [+0.11, +0.45] & +1.25 [+1.11, +1.41] \\
  gpt-oss-20b (Bedrock) & +0.77 [+0.60, +0.93] & +0.83 [+0.69, +0.98] & +0.97 [+0.85, +1.10] \\
  gpt-oss-120b (Bedrock) & +0.97 [+0.83, +1.12] & +0.94 [+0.80, +1.08] & +1.08 [+0.97, +1.20] \\
 
    \bottomrule
  \end{tabular}
  \end{adjustbox}

  \vspace{6pt}
  \begin{adjustbox}{max width=\linewidth}
  \begin{tabular}{@{}lrrrrr@{}}
    \toprule
    Model & C4 & C5 & \cnd{fewer\_ok} vs \cnd{none} & Unreadable \% & Missing \\
    \midrule
      DeepSeek V4.1 Flash (Together) & +0.54 [+0.49, +0.59] & $-$0.64 [$-$0.71, $-$0.57] & +0.53 [+0.39, +0.67] & 0.0 & 0 \\
  Qwen3.5-9B (Together) & +0.08 [+0.04, +0.13] & $-$0.58 [$-$0.67, $-$0.51] & $-$0.61 [$-$0.75, $-$0.48] & 0.0 & 0 \\
  Kimi K3 (Bedrock) & +0.55 [+0.49, +0.60] & -- & +0.32 [+0.24, +0.40] & 0.0 & 0 \\
  DeepSeek V3.2 (Bedrock) & +0.33 [+0.29, +0.37] & -- & +0.36 [+0.22, +0.50] & 0.1 & 0 \\
  Qwen3-235B (Bedrock) & +0.30 [+0.26, +0.34] & -- & +0.30 [+0.16, +0.43] & 0.0 & 0 \\
  Mistral Large 3 (Bedrock) & +0.30 [+0.26, +0.34] & -- & $-$0.49 [$-$0.63, $-$0.35] & 0.3 & 0 \\
  Ministral 3 14B (Bedrock) & +0.22 [+0.17, +0.26] & -- & $-$1.59 [$-$1.84, $-$1.34] & 0.1 & 0 \\
  gpt-oss-20b (Bedrock) & +0.27 [+0.21, +0.32] & -- & +0.07 [$-$0.06, +0.20] & 0.5 & 0 \\
  gpt-oss-120b (Bedrock) & +0.50 [+0.45, +0.54] & -- & +0.01 [$-$0.11, +0.12] & 0.0 & 0 \\
 
    \bottomrule
  \end{tabular}
  \end{adjustbox}
  \caption{Hosted reference models, never pooled with the GPU models; contrasts as in Table~\ref{tab:permodel}. Bedrock runs the prompt-side conditions only, so contrast 5 is empty there.}
  \label{tab:hosted}
\end{table*}

\begin{table}[t]
  \centering
  \small
  \setlength{\tabcolsep}{3pt}
  \begin{adjustbox}{max width=\linewidth}
  \begin{tabular}{@{}lrrrrr@{}}
    \toprule
    Condition & Outputs & Padded & Says short & Short, no note & Unread. \\
    \midrule
      \cnd{none} & 8181 & 34.8 & 0.0 & 65.2 & 0.0 \\
  \cnd{prose\_up\_to\_k} & 40905 & 44.0 & 0.0 & 56.0 & 0.1 \\
  \cnd{prose\_exactly\_k} & 40905 & 56.2 & 0.2 & 43.5 & 0.1 \\
  \cnd{max\_k\_shown} & 8181 & 38.8 & 0.0 & 61.1 & 0.1 \\
  \cnd{min\_k\_shown} & 8181 & 65.8 & 2.4 & 30.9 & 1.0 \\
  \cnd{min\_k\_shown\_fewer\_ok} & 8181 & 38.5 & 1.0 & 59.9 & 0.6 \\
  \cnd{min\_k\_enforced} & 1818 & 90.7 & 3.0 & 6.2 & 0.1 \\
  \cnd{min\_k\_escape} & 1818 & 49.9 & 49.7 & 0.3 & 0.0 \\
 
    \bottomrule
  \end{tabular}
  \end{adjustbox}
  \caption{Sorting, hosted models pooled (\%), $n < k$, $k = 3$ and 5. The enforced conditions are Together only.}
  \label{tab:hostedsorting}
\end{table}

\begin{table*}[t]
  \centering
  \small
  \begin{adjustbox}{max width=\linewidth}
  \begin{tabular}{@{}lrrrrl@{}}
    \toprule
    Pair & Hosted & GPU & Difference & Difference (pts) & Under 10 pts \\
    \midrule
      Qwen3.5-9B (Together) vs Qwen3.5-9B & +0.64 [+0.50, +0.78] & +0.62 [+0.50, +0.74] & +0.02 [$-$0.09, +0.13] & +0.7 [$-$1.3, +2.6] & yes \\
  Ministral 3 14B (Bedrock) vs Ministral 3 14B & +0.28 [+0.11, +0.45] & +0.30 [+0.12, +0.45] & $-$0.02 [$-$0.16, +0.12] & $-$1.1 [$-$3.4, +1.1] & yes \\
  gpt-oss-20b (Bedrock) vs gpt-oss-20b & +0.83 [+0.69, +0.98] & +0.87 [+0.72, +1.01] & $-$0.04 [$-$0.18, +0.11] & +0.8 [$-$2.3, +3.8] & yes \\
 
    \bottomrule
  \end{tabular}
  \end{adjustbox}
  \caption{Same-weights pairs: contrast 2 in entries on the same documents and $k$, and the difference in over-count points, which the registered 10-point rule is judged on.}
  \label{tab:pairs}
\end{table*}

\section{Repeat check}
\label{app:repeat}

Table~\ref{tab:repeat} gives, per model, how often the same prompt sent twice gave the same reply.

\begin{table}[t]
  \centering
  \small
  \setlength{\tabcolsep}{3pt}
  \begin{adjustbox}{max width=\linewidth}\begin{tabular}{lrrrr}
  \toprule
  Model & Prompts & Identical \% & Flips \% & Length differs \% \\
  \midrule
  DeepSeek V3.2 (Bedrock) & 300 & 52.7 & 4.00 & 12.0 \\
  gpt-oss-120b (Bedrock) & 300 & 18.0 & 8.00 & 13.0 \\
  gpt-oss-20b (Bedrock) & 300 & 31.0 & 4.00 & 10.3 \\
  Kimi K3 (Bedrock) & 300 & 92.7 & 0.67 & 2.7 \\
  Ministral 3 14B (Bedrock) & 300 & 67.0 & 1.33 & 5.3 \\
  Mistral Large 3 (Bedrock) & 300 & 54.0 & 7.00 & 14.7 \\
  Qwen3-235B (Bedrock) & 300 & 64.3 & 3.00 & 4.3 \\
  Gemma 4 31B & 300 & 98.7 & 0.00 & 0.3 \\
  gpt-oss-20b & 300 & 59.3 & 3.00 & 11.3 \\
  Ministral 3 14B & 300 & 88.7 & 1.00 & 3.0 \\
  Olmo 3.1 32B & 300 & 87.7 & 0.67 & 2.3 \\
  Qwen3.5-0.8B & 300 & 71.7 & 3.00 & 12.3 \\
  Qwen3.5-27B & 300 & 94.7 & 0.33 & 0.7 \\
  Qwen3.5-2B & 300 & 82.3 & 1.67 & 9.3 \\
  Qwen3.5-9B & 300 & 90.3 & 0.67 & 2.0 \\
  DeepSeek V4.1 Flash (Together) & 300 & 65.3 & 5.00 & 9.7 \\
  Qwen3.5-9B (Together) & 300 & 84.0 & 2.33 & 5.7 \\
  \bottomrule
\end{tabular}
\end{adjustbox}
  \caption{300 prompts per model sent twice (for GPU models, the repeat on a newly started server): identical replies, over-count flips and a changed number of substantive entries (\%).}
  \label{tab:repeat}
\end{table}

\section{Exploratory analyses}
\label{app:reask}

The re-ask arm (Table~\ref{tab:reask}) and the protocol's research questions on asking for fewer items than exist (RQ3, Table~\ref{tab:rq3}) and on item type (RQ4) are exploratory.

\begin{table*}[t]
  \centering
  \small
  \begin{adjustbox}{max width=\linewidth}
  \begin{tabular}{@{}lrrrrrrl@{}}
    \toprule
    Model & Re-asked & $\geq k$ after \% & With $n < k$ & Over before \% & Over after \% & Length change & Sorting after \\
    \midrule
      Kimi K3 (Bedrock) & 510 & 100.0 & 508 & 14.0 & 78.3 & +1.89 & 78 / 6 / 16 / 0 \\
  Ministral 3 14B (Bedrock) & 410 & 85.6 & 405 & 21.7 & 81.5 & +1.99 & 81 / 10 / 3 / 6 \\
  Gemma 4 31B & 429 & 89.7 & 427 & 11.2 & 44.0 & +0.63 & 44 / 30 / 19 / 7 \\
  Ministral 3 14B & 425 & 86.6 & 419 & 21.5 & 81.6 & +1.98 & 82 / 9 / 4 / 5 \\
  Qwen3.5-9B & 678 & 73.0 & 639 & 10.0 & 69.0 & +1.70 & 69 / 1 / 30 / 0 \\
 
    \bottomrule
  \end{tabular}
  \end{adjustbox}
  \caption{Re-ask arm: \cnd{min\_k\_shown} readable replies shorter than $k$, at every $k$, sent back once with the validation error. Over-count and length change are on the re-asked replies with $n < k$; $\geq k$ after is over all re-asked replies. Sorting after, on $n < k$: padded / says short / short, no note / unreadable (\%).}
  \label{tab:reask}
\end{table*}

\begin{table*}[t]
  \centering
  \small
  \begin{adjustbox}{max width=\linewidth}
  \begin{tabular}{@{}lrrrrrr@{}}
    \toprule
    Condition & 1st & 2nd & 3rd & 4th & 5th & 6th+ \\
    \midrule
      \cnd{prose\_up\_to\_k} & 28.6 & 45.0 & 47.9 & 48.5 & 56.4 & 49.4 \\
  \cnd{prose\_exactly\_k} & 29.7 & 53.4 & 56.4 & 57.7 & 64.4 & 62.0 \\
  \cnd{max\_k\_shown} & 25.9 & 36.1 & 37.5 & 41.7 & 47.9 & 48.5 \\
  \cnd{min\_k\_shown} & 19.1 & 18.1 & 17.1 & 16.4 & 19.1 & 14.0 \\
  \cnd{min\_k\_shown\_fewer\_ok} & 21.1 & 19.9 & 19.7 & 19.4 & 23.3 & 19.3 \\
  \cnd{min\_k\_enforced} & 13.6 & 11.2 & 10.6 & 8.6 & 13.8 & 7.2 \\
  \cnd{min\_k\_escape} & 16.6 & 13.9 & 13.8 & 11.8 & 15.3 & 10.2 \\
 
    \bottomrule
  \end{tabular}
  \end{adjustbox}
  \caption{RQ3, asked for fewer than exist ($k < n$): share of true items left out\pre{} by position of first appearance in the document (\%), eight GPU models pooled.}
  \label{tab:rq3}
\end{table*}

RQ4 (item type inside job ads): contrast 2 is $+1.54$ [0.83, 2.30] entries for skills (quoted) and $+0.76$ [$-0.26$, $+1.79$] for knowledge (named tools); skills minus knowledge is $+0.78$ [$-0.48$, $+2.05$].
RQ5 (probability of closing the list), RQ7 (token confidence of extra entries) and RQ8 (hidden-state probe) are not reported: they need the teacher-forced and hidden-state passes, which were not run on the full-study outputs, and the robustness arms (LlamaIndex's schema block, a neutral instruction) ran in the pilot only (Table~\ref{tab:deviations}).

\section{Thinking side test}
\label{app:thinking}

Each pair differs in the thinking switch: Qwen3.5-9B and Gemma 4 31B run with thinking on and off under the vendor's sampling, a Qwen3.5-9B pair with the presence penalty at 0 runs on the 60 pilot documents, and Kimi K3 runs with thinking on in the prompt-side conditions on the pilot documents, against its thinking-off full run on the same documents, which is greedy, so that pair also differs in temperature (1.0 against 0; Table~\ref{tab:thinking}).

\begin{table*}[t]
  \centering
  \small
  \begin{adjustbox}{max width=\linewidth}
  \begin{tabular}{@{}llrrrrrr@{}}
    \toprule
    & & \multicolumn{3}{c}{C2 (entries)} & \multicolumn{2}{c}{Over-count (\%)} & Names \\
    \cmidrule(lr){3-5} \cmidrule(lr){6-7}
    Pair & Seed & on & off & on $-$ off & on & off & shortfall (\%) \\
    \midrule
      Qwen3.5-9B & 1 & +1.77 [+1.62, +1.92] & +0.64 [+0.50, +0.77] & +1.14 [+0.94, +1.32] & 81.3 & 45.2 & 87.9 \\
   & 2 & +1.78 [+1.64, +1.92] & +0.59 [+0.44, +0.74] & +1.19 [+0.98, +1.39] & 83.2 & 47.1 & 89.0 \\
   & 3 & +1.69 [+1.46, +1.88] & +0.55 [+0.38, +0.71] & +1.14 [+0.91, +1.36] & 81.5 & 43.0 & 88.3 \\
  Qwen3.5-9B, penalty 0 & 1 & +2.45 [+2.16, +2.73] & +1.37 [+1.03, +1.72] & +1.08 [+0.63, +1.51] & 100.0 & 75.0 & 80.1 \\
   & 2 & +2.33 [+2.00, +2.64] & +1.25 [+0.69, +1.71] & +1.08 [+0.62, +1.56] & 99.2 & 80.0 & 84.5 \\
   & 3 & +2.43 [+2.09, +2.78] & +1.29 [+0.91, +1.66] & +1.14 [+0.68, +1.62] & 98.3 & 74.2 & 82.4 \\
  Gemma 4 31B & 1 & +1.52 [+1.39, +1.64] & +1.63 [+1.50, +1.76] & $-$0.11 [$-$0.23, +0.01] & 66.1 & 61.8 & 59.2 \\
  Kimi K3 (Bedrock) & - & +2.42 [+2.09, +2.76] & +1.84 [+1.48, +2.24] & +0.58 [+0.23, +0.91] & 97.5 & 91.7 & 29.6 \\
 
    \bottomrule
  \end{tabular}
  \end{adjustbox}
  \caption{Thinking on against off, documents with $n < k$, $k = 3$ and 5. C2 is the length change of \cnd{min\_k\_shown} over \cnd{none} in substantive entries, with 95\% intervals from the group bootstrap; over-count is under \cnd{min\_k\_shown}. The last column is the share of over-counting outputs with thinking on ($n < k$, every condition with a count) whose thinking text names the shortfall, by a phrase search. Sampled decoding (Kimi K3 with thinking off: greedy), one row per request seed; the Kimi K3 shortfall share also counts $k = 1$. The presence-penalty-0 pair and Kimi K3 run on the 60 pilot documents only.}
  \label{tab:thinking}
\end{table*}

\section{Post hoc analyses}
\label{app:posthoc}

None of these analyses is in the registered plan. Unless stated, each uses the stored full-study outputs with the frozen parser and matcher unchanged, on the eight GPU models pooled.

\paragraph{Source check.}
A developer can check that each extracted entry occurs in its source document. We test two versions: the entry, normalised for case, quotes, dashes and spaces, as an exact substring of the normalised document, and a fuzzy match (partial ratio of at least 90).
The check removes invented entries: 0.6\% of the matcher's invented entries pass, and none of the 99 entries the authors labelled invented passes either version.
It lets most grounded padding through (67.7\% exact) and about half of all extra entries (46.6\% under a shown minimum), against 80.6\% of true-item entries (Table~\ref{tab:sourcecheck}).
Meetings sit apart: their items are one-sentence summaries, not copies, so only 7.5\% of their extra entries pass.

\begin{table}[t]
  \centering
  \small
  \setlength{\tabcolsep}{3pt}
  \begin{adjustbox}{max width=\linewidth}
  \begin{tabular}{@{}lrrr@{}}
    \toprule
    Entries & Count & Exact \% & Fuzzy \% \\
    \midrule
      \cnd{none} & 18,109 & 46.5 & 49.0 \\
  \cnd{prose\_up\_to\_k} & 62,574 & 48.4 & 50.7 \\
  \cnd{prose\_exactly\_k} & 66,808 & 53.7 & 56.0 \\
  \cnd{max\_k\_shown} & 14,697 & 47.3 & 49.8 \\
  \cnd{min\_k\_shown} & 22,665 & 46.6 & 48.7 \\
  \cnd{min\_k\_shown\_fewer\_ok} & 13,343 & 56.5 & 58.7 \\
  \cnd{min\_k\_enforced} & 26,193 & 65.6 & 68.2 \\
  \cnd{min\_k\_escape} & 15,198 & 67.6 & 70.6 \\
  \midrule
  true-item entries (reference) & 137,849 & 80.6 & 83.2 \\
  extra: wrong kind & 173,345 & 67.7 & 70.7 \\
  extra: split true item & 18,025 & 52.4 & 54.6 \\
  extra: invented & 48,217 & 0.6 & 0.9 \\
 
    \bottomrule
  \end{tabular}
  \end{adjustbox}
  \caption{Post hoc source check: share of entries that occur in their source document, eight GPU models pooled, $n < k$, $k = 3$ and 5. Top: extra entries by condition. Bottom: all conditions, by matcher label\pre{}, with true-item entries for reference.}
  \label{tab:sourcecheck}
\end{table}

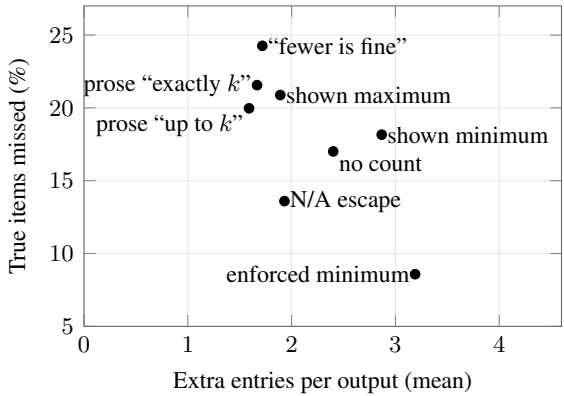
\begin{figure}[t]
  \centering
  \begin{tikzpicture}
\begin{axis}[scale only axis, width=0.82\columnwidth, height=0.55\columnwidth,
  xmin=0, xmax=4.6, ymin=5, ymax=27, grid=major, grid style={gray!20}, axis line style={black!60},
  xlabel={Extra entries per output (mean)}, ylabel={True items missed (\%)},
  label style={font=\footnotesize}, tick label style={font=\footnotesize}]
\addplot[only marks, mark=*, mark size=1.8pt, color=black] coordinates {(2.401,17.01) (1.590,19.97) (1.667,21.56) (1.891,20.88) (2.868,18.16) (1.718,24.26) (3.190,8.58) (1.931,13.60)};
\node[font=\footnotesize, anchor=north west, inner sep=2pt] at (axis cs:2.401,17.01) {\cnd{none}};
\node[font=\footnotesize, anchor=north east, inner sep=2pt] at (axis cs:1.590,19.97) {\cnd{prose\_up\_to\_k}};
\node[font=\footnotesize, anchor=east, inner sep=2pt] at (axis cs:1.667,21.56) {\cnd{prose\_exactly\_k}};
\node[font=\footnotesize, anchor=west, inner sep=2pt] at (axis cs:1.891,20.88) {\cnd{max\_k\_shown}};
\node[font=\footnotesize, anchor=west, inner sep=2pt] at (axis cs:2.868,18.16) {\cnd{min\_k\_shown}};
\node[font=\footnotesize, anchor=west, inner sep=2pt] at (axis cs:1.718,24.26) {\cnd{min\_k\_shown\_fewer\_ok}};
\node[font=\footnotesize, anchor=east, inner sep=2pt] at (axis cs:3.190,8.58) {\cnd{min\_k\_enforced}};
\node[font=\footnotesize, anchor=west, inner sep=2pt] at (axis cs:1.931,13.60) {\cnd{min\_k\_escape}};
\end{axis}
\end{tikzpicture}
  \caption{Post hoc: mean extra entries per output against the share of true items missed\pre{}, one point per condition, eight GPU models pooled, all documents, $k = 3$ and 5 (the rows of Table~\ref{tab:prf}). Per-model points are in the repository.}
  \label{fig:padmiss}
\end{figure}

\paragraph{Precision and recall.}
Over all documents, a shown minimum lowers entry precision from 0.39 with no count to 0.35 and leaves recall about where it was (0.83 and 0.82); an enforced minimum raises recall to 0.91 at the same precision, because it adds the most entries (Table~\ref{tab:prf}).
Prose ``up to $k$'' has the best F1 (0.60) and the N/A escape the next best (0.59), against 0.53 with no count.
Figure~\ref{fig:padmiss} places each condition by its padding and the share of true items it misses.

\begin{table}[t]
  \centering
  \small
  \setlength{\tabcolsep}{3pt}
  \begin{adjustbox}{max width=\linewidth}
  \begin{tabular}{@{}lrrrr@{}}
    \toprule
    Condition & Precision & Recall & F1 & Extra / output \\
    \midrule
      \cnd{none} & 0.39 & 0.83 & 0.53 & 2.40 \\
  \cnd{prose\_up\_to\_k} & 0.48 & 0.80 & 0.60 & 1.59 \\
  \cnd{prose\_exactly\_k} & 0.46 & 0.78 & 0.58 & 1.67 \\
  \cnd{max\_k\_shown} & 0.43 & 0.79 & 0.56 & 1.89 \\
  \cnd{min\_k\_shown} & 0.35 & 0.82 & 0.49 & 2.87 \\
  \cnd{min\_k\_shown\_fewer\_ok} & 0.45 & 0.76 & 0.56 & 1.72 \\
  \cnd{min\_k\_enforced} & 0.35 & 0.91 & 0.50 & 3.19 \\
  \cnd{min\_k\_escape} & 0.45 & 0.86 & 0.59 & 1.93 \\
 
    \bottomrule
  \end{tabular}
  \end{adjustbox}
  \caption{Post hoc entry-level precision, recall and F1\pre{} against the true items, eight GPU models pooled, all documents, $k = 3$ and 5. Precision is over all substantive entries, recall over the true items of documents with $n \geq 1$; unreadable outputs count as empty lists. Intervals and per-model values are in the repository.}
  \label{tab:prf}
\end{table}

\paragraph{Qwen3.5-0.8B read leniently.}
A lenient parser that also takes the list from inside a copy of the schema, and changes none of the outputs the frozen parser already reads, cuts this model's unreadable share from 36.6\% to 9.8\%.
Read this way, it over-counts on 52.8\% of its $n < k$ outputs with no count and on 51.5\% under a shown minimum ($-1.3$ points [$-4.0$, $+1.4$]); on the 622 pairs readable under both conditions, a shown minimum adds 7.4 over-count points [5.0, 9.9] and $+0.23$ entries [$-0.36$, $+0.80$].
Its negative estimate in Table~\ref{tab:permodel} comes from unreadable answers counted as empty lists, not from shorter lists.
Without this model, the seven other GPU models give a main contrast of $+0.80$ entries [0.63, 0.96] against the registered $+0.62$, $+24.7$ over-count points [23.1, 26.4] against $+20.9$, and over-count shares of 35.0\% with no count, 59.8\% under a shown minimum and 89.9\% under an enforced one (\texttt{analysis/posthoc/without\_08b.py}).

\paragraph{How much of the padding is invented.}
Under a shown minimum, on documents with $n < k$, the matcher calls 29.5\% of the GPU models' extra entries invented; spread over the authors' labels in the proportions of the 400-entry check, the share is 23.4\% [20.4, 26.6].
The 7.0\% and 7.3\% of Appendix~\ref{app:checks} are shares of all substantive entries, true items included, over every model and condition.
On the same basis, the hosted models' extra entries are 5.6\% invented.

\paragraph{Documents a model gets right with no count.}
The 34.5\% over-count with no count mixes ordinary extraction error into every absolute rate.
To set it aside, we keep only the (model, document) pairs whose no-count list has exactly $n$ substantive entries (2,275 pairs, 3,884 cells of $k = 3$ or 5), or at most $n$ (4,762 cells), so the no-count over-count is 0 by construction.
On these pairs a shown minimum makes 35.2\% of lists pad, a shown maximum 5.0\%, ``fewer is fine'' 9.6\% and an enforced minimum 83.9\% (Table~\ref{tab:cleanbase}).
Under a shown maximum, which changes the prompt as much, 5.0\% pad, so most of the 35.2\% comes from the minimum itself.
Without Qwen3.5-0.8B the shown-minimum share is 36.7\% (\texttt{analysis/posthoc/clean\_baseline.py}).

\begin{table}[t]
  \centering
  \small
  \setlength{\tabcolsep}{3pt}
  \begin{adjustbox}{max width=\linewidth}
  \begin{tabular}{@{}lrr@{}}
    \toprule
    Condition & No count exactly $n$ & No count at most $n$ \\
    \midrule
      \cnd{min\_k\_shown} & 35.2 [33.0, 37.5] & 33.9 [32.1, 35.8] \\
  \cnd{min\_k\_shown\_fewer\_ok} & 9.6 [8.5, 10.7] & 10.1 [9.0, 11.2] \\
  \cnd{min\_k\_enforced} & 83.9 [82.2, 85.5] & 85.3 [83.8, 86.7] \\
  \cnd{min\_k\_escape} & 29.2 [27.2, 31.2] & 36.1 [34.0, 38.1] \\
  \cnd{max\_k\_shown} & 5.0 [4.2, 5.9] & 5.1 [4.3, 5.8] \\
 
    \bottomrule
  \end{tabular}
  \end{adjustbox}
  \caption{Post hoc: over-count (\%, 95\% group-bootstrap interval) on the (model, document) pairs whose no-count list was exactly $n$ entries long or at most $n$, eight GPU models pooled, $n < k$, $k = 3$ and 5. With no count the rate is 0 by construction.}
  \label{tab:cleanbase}
\end{table}

\paragraph{How often real schemas set a list minimum.}
We count list minimums in the 9,542 real-world schemas of JSONSchemaBench \citep{geng2025jsonschemabench} (MIT licence, revision \texttt{5bd0f46}), leaving out fixed-length arrays and tuples such as coordinates (Table~\ref{tab:schemaprev}).
Half the schemas hold an array; 7.7\% put a \texttt{minItems} of at least 1 on a list of strings or objects, and 0.3\% (24 schemas) one of 2 or more, such as a list of players or of interests.
The common schema minimum is therefore 1, the case of the empty-document test (Section~\ref{sec:named}); larger minimums are rare in public schemas.
The function-calling subset (Glaiveai2K) sets none (\texttt{analysis/posthoc/schema\_prevalence.py}).

\begin{table}[t]
  \centering
  \small
  \setlength{\tabcolsep}{3pt}
  \begin{adjustbox}{max width=\linewidth}
  \begin{tabular}{@{}lrrrrr@{}}
    \toprule
    Subset & Schemas & Array \% & Min $\geq 1$, list \% & Min $\geq 2$ \% & Min $\geq 2$, list \% \\
    \midrule
      Github\_trivial & 444 & 17.6 & 1.8 & 0.9 & 0.0 \\
  Github\_easy & 1,938 & 33.4 & 4.8 & 0.7 & 0.4 \\
  Github\_medium & 1,969 & 57.4 & 9.8 & 0.7 & 0.3 \\
  Github\_hard & 1,236 & 78.8 & 22.7 & 0.7 & 0.5 \\
  Github\_ultra & 164 & 95.7 & 36.0 & 3.0 & 1.8 \\
  Glaiveai2K & 1,707 & 29.1 & 0.0 & 0.0 & 0.0 \\
  JsonSchemaStore & 492 & 88.8 & 15.0 & 0.8 & 0.2 \\
  Kubernetes & 1,064 & 60.8 & 0.0 & 0.0 & 0.0 \\
  Snowplow & 403 & 33.7 & 6.0 & 0.0 & 0.0 \\
  WashingtonPost & 125 & 48.8 & 4.0 & 0.0 & 0.0 \\
  \midrule
  All & 9,542 & 49.9 & 7.7 & 0.5 & 0.3 \\
 
    \bottomrule
  \end{tabular}
  \end{adjustbox}
  \caption{Post hoc: list minimums in JSONSchemaBench, share of schemas. Array: any array. Min $\geq 1$, list: an open array of strings or objects with \texttt{minItems} $\geq 1$. Min $\geq 2$: any open array with \texttt{minItems} $\geq 2$; list: of strings or objects. Open arrays exclude fixed-length arrays (\texttt{minItems} = \texttt{maxItems}) and tuples.}
  \label{tab:schemaprev}
\end{table}

\paragraph{Examples.}
One over-counting output per set under a shown minimum, on a document with $1 \leq n < k$ and with at most six short entries, chosen by a fixed hash rule, not by hand.
The labels are the matcher's; in the emails example it swaps entries 1 and 2, though the number of extra entries is right.

\paragraph{Meetings.} Olmo 3.1 32B, $k = 3$, $n = 1$: 3 entries, 2 extra.
Asked for: Next steps that a meeting participant will take after this meeting, one per job. One short sentence each. Not decisions, and not instructions they will receive.
True items: ``The group will find out if the kinetic energy component can provide enough energy.''.
Output: (1) ``Check with the R\&D department to see if the kinetic metal technology can provide sufficient energy for the remote.'' [true item]; (2) ``Consult with the R\&D department regarding the readiness of the kinetic energy product for market.'' [split true item, extra]; (3) ``Gather and share pictures of the device cases with the Industrial Designer for material identification (titanium vs. plastic).'' [wrong kind, extra].

\paragraph{Emails.} Olmo 3.1 32B, $k = 3$, $n = 2$: 4 entries, 2 extra.
Asked for: Things the sender asks the recipient to do or to provide, one per thing asked. Copy each request as written in the email.
True items: ``What do we need to do to get this''; ``how soon can it be granted?''.
Output: (1) ``Everyone in risk management needs to get access to EOL bridge monitor and reports.'' [true item]; (2) ``What do we need to do to get this access?'' [split true item, extra]; (3) ``How soon can the access be granted?'' [true item]; (4) ``We need this access by Thursday to be prepared for testing.'' [wrong kind, extra].

\paragraph{Receipts.} Qwen3.5-9B, $k = 3$, $n = 1$: 3 entries, 2 extra.
Asked for: Names of the items purchased on this receipt, one entry per item, copied as printed.
True items: ``ALMOND CHOCO CREAM CHEESE''.
Output: (1) ``ALMOND CHOCO CREAM CHEESE'' [true item]; (2) ``SUB TOTAL'' [wrong kind, extra]; (3) ``GRAND TOTAL'' [wrong kind, extra].

\paragraph{Job-ad skills.} Olmo 3.1 32B, $k = 5$, $n = 1$: 5 entries, 4 extra.
Asked for: Skills, abilities and duties the job ad mentions, copied as written. Not tools, technologies, fields of knowledge, languages or certificates.
True items: ``running with a full stack development''.
Output: (1) ``running full stack development'' [true item]; (2) ``Java'' [wrong kind, extra]; (3) ``Python'' [wrong kind, extra]; (4) ``React JS'' [wrong kind, extra]; (5) ``Redux'' [wrong kind, extra].

\paragraph{Clinical.} Ministral 3 14B, $k = 5$, $n = 3$: 5 entries, 2 extra.
Asked for: Medications or drugs, by name or by class, mentioned as given to or taken by the patient, copied as written.
True items: ``ACE inhibitor''; ``diuretic''; ``$\beta$-blocker''.
Output: (1) ``ACE inhibitor'' [true item]; (2) ``diuretic'' [true item]; (3) ``low-dose $\beta$-blocker'' [true item]; (4) ``coronary angiography contrast agent (if applicable)'' [invented, extra]; (5) ``endomyocardial biopsy procedural sedation/anesthesia (if applicable)'' [invented, extra].

\paragraph{Reviews.} Qwen3.5-2B, $k = 5$, $n = 1$: 5 entries, 4 extra.
Asked for: Features or aspects of the product or service that the reviewer gives an opinion on, copied as written.
A short hotel review from the restricted reviews set (no review text is reproduced here). The reviewer calls the hotel a hidden gem of its city, says they had expected an ordinary chain hotel, says it beat every hotel they had used in a long time, and praises the staff. The answer list holds one item: the staff.

The model's list has five entries. Four copy short phrases from the review that name the hotel or praise it in general (its headline, the chain it was expected to resemble, the comparison with other hotels); the matcher calls all four real text of the wrong kind, so all four are extra. The fifth, praise of the staff, is the one true item. (Paraphrased: review text is not redistributed.)

\paragraph{Job-ad knowledge.} Gemma 4 31B, $k = 5$, $n = 2$: 5 entries, 3 extra.
Asked for: Tools, technologies, fields of knowledge, languages and certificates the job ad mentions, copied as written. Not abilities or duties.
True items: ``science''; ``pharmaceutical production''.
Output: (1) ``science education'' [true item]; (2) ``pharmaceutical production'' [true item]; (3) ``management experience'' [wrong kind, extra]; (4) ``communication skills'' [wrong kind, extra]; (5) ``collaboration skills'' [wrong kind, extra].

\paragraph{Wikipedia people.} Qwen3.5-2B, $k = 5$, $n = 2$: 3 entries, 1 extra.
Asked for: Names of people mentioned in the text, copied as written.
True items: ``Stamatiad''; ``Emil Gârleanu''.
Output: (1) ``Stamatiad'' [true item]; (2) ``Emil Gârleanu'' [true item]; (3) ``Romanian Writers ' Society (SSR)'' [wrong kind, extra].

\paragraph{Novel quotes.} gpt-oss-20b, $k = 5$, $n = 3$: 5 entries, 2 extra.
Asked for: Direct speech quoted in the text, copied word for word.
True items: ``After what happened at Vevey, I think I may certainly call upon them,''; ``If, after what happens-{}-at Vevey and everywhere-{}-you desire to keep up the acquaintance, you are very welcome. Of course a man may know everyone. Men are welcome to the privilege!''; ``Pray what is it that happens-{}-here, for instance?''.
Output: (1) ``After what happened at Vevey'' [true item]; (2) ``I think I may certainly call upon them,'' [split true item, extra]; (3) ``If, after what happens-{}-at Vevey and everywhere-{}-you desire to keep up the acquaintance, you are very welcome.'' [true item]; (4) ``Of course a man may know everyone. Men are welcome to the privilege!'' [split true item, extra]; (5) ``Pray what is it that happens-{}-here, for instance?'' [true item].

\section{Pilot stop rules}
\label{app:stoprules}

Both pilot stop rules on over-count say go: on pilot documents, $n < k$, $k = 3$ and 5, pooled over the three pilot models, \cnd{min\_k\_shown} minus \cnd{none} was $+34.4$ points [28.1, 40.8] (Qwen3.5-9B 30.0, Ministral 3 14B 13.3, Gemma 4 31B 60.0) and \cnd{min\_k\_shown} minus \cnd{max\_k\_shown} $+34.2$ [28.3, 40.3].
The third stop rule, the blind check at the registered threshold of 85\%, passes (Appendix~\ref{app:key}).
The stop-rule simulation, rerun with the pilot's floor rates (600 simulations), stops at a true gap of 0 points with probability 0.68 to 0.80 and goes at a true gap of 10 points with probability 0.67 to 0.82 across floors of 15 to 78\%.

\end{document}